\documentclass[a4paper,12pt]{elsarticle}

\usepackage[utf8]{inputenc}
\usepackage[usenames,dvipsnames,svgnames]{xcolor}
\usepackage{color,soul}
\usepackage{amsmath,amssymb,amsthm}
\usepackage{booktabs}
\usepackage{adjustbox}
\usepackage[colorinlistoftodos,prependcaption,textsize=footnotesize]{todonotes}
\usepackage{placeins}
\usepackage{subcaption} 
\usepackage{multirow}
\usepackage{tabularx}
\usepackage{mathabx}
\usepackage{hyperref}

\definecolor{mypink1}{rgb}{0.958, 0.688, 0.78}

\newif\ifmargincomments
\margincommentsfalse

\newif\ifreviewmarkings
\reviewmarkingsfalse

\ifmargincomments

\else

\fi

\ifreviewmarkings
\newcommand{\revI}[1]{{\color{black}#1}}
\newcommand{\revII}[1]{{\color{blue}#1}}
\else
\newcommand{\revI}[1]{#1}
\newcommand{\revII}[1]{#1}
\fi

\graphicspath{
{./figs/}
{./}
}

\title{Proximal Exploration of Venus Volcanism with Teams of Autonomous Buoyancy-Controlled Balloons}

\author[1]{Federico Rossi\corref{cor1}}
\ead{Federico.Rossi@jpl.nasa.gov}
\author[1]{Maíra Saboia}
\ead{mairasab@buffalo.edu}
\author[1]{Siddharth Krishnamoorthy}
\ead{Siddharth.Krishnamoorthy@jpl.nasa.gov}
\author[1]{Joshua Vander Hook}
\ead{hook@jpl.nasa.gov}
\address[1]{Jet Propulsion Laboratory, California Institute of Technology, Pasadena, CA, USA}

\cortext[cor1]{Corresponding author}

\begin{document}

\begin{abstract}

Altitude-controlled balloons hold great promise for performing high-priority scientific investigations of Venus's atmosphere and geological phenomena, including tectonic and volcanic activity, as demonstrated by a number of recent Earth-based experiments.

In this paper, we explore a concept of operations where multiple autonomous, altitude-controlled balloons monitor \revI{explosive} volcanic activity on Venus through infrasound microbarometers, and autonomously navigate the uncertain wind field  to perform follow-on observations of detected events of interest. We propose a novel autonomous guidance technique for altitude-controlled balloons in Venus's uncertain wind field, and show the approach can result in an increase of up to 63\% in the number of close-up observations of volcanic events compared to passive drifters, and a 16\% increase compared to ground-in-the-loop guidance. The results are robust to uncertainty in the wind field, and hold across large changes in the frequency of \revI{explosive} volcanic events, sensitivity of the microbarometer detectors, and numbers of aerial platforms. 
\end{abstract}

\maketitle
\textcopyright \the\year{}, California Institute of Technology. Government sponsorship acknowledged.

\section{Introduction}

Venus exploration has seen renewed interest from the planetary science community in the last decade, especially since the as-yet contested discovery of phosphine gas \cite{greaves2021phosphine,villanueva2021no}, a potential biosignature, in its atmosphere. Venus' physical similarity to Earth (0.81 Earth mass and 0.86 Earth volume), yet stark differences in atmospheric composition, climate, and topography continue to remain a mystery. NASA recently selected \revI{VERITAS (Venus Emissivity, Radio Science, InSAR, Topography, and Spectroscopy) \cite{smrekar2022veritas} and  DAVINCI+ (Deep Atmosphere Venus Investigation of Noble gases, Chemistry, and Imaging)\revI{\cite{getty2021davinci,garvin2022revealing}}} as Discovery class missions to Venus, while the European \revI{Space Agency} selected the EnVision mission \cite{ghail2012envision} as \revI{the fifth medium mission (M5) of the Cosmic Vision program}. Several other nations and private entities have also announced \revI{potential} missions to Venus in the next decade \cite{oCallaghan2020life,Haider2018Shukrayaan,Zasova2019VeneraD,french2021bringing}. However, despite over 50 years of exploration, the interior structure of Venus remains a mystery, hampering our understanding of Venus's evolution as a planet, and its divergent path from Earth. The characterization of current volcanic and tectonic activity is therefore considered an essential science investigation by the Venus exploration community~\cite{vexag2019goi}.

In the 1960s and 70s, several landed, balloon-based, and orbital missions explored Venus, revealed much of what we know of the planet today. Images from the surface of Venus returned by the Venera landers revealed a planet that was predominantly shaped by volcanism (e.~g.,~\cite{vinogradov1973venera}). \revI{Radar data from the Venera, Pioneer, and Magellan spacecraft revealed a surface covered with a plethora of volcanic structures such as cones, domes, and coronae \cite{head1992venus}.} In addition, even though large tectonic plates are thought to be absent on Venus, several signs of geologically recent seismic strain at a local scale are visible in Magellan data \cite{byrne2021litho}. Several studies conducted using data from Venus missions \cite{esposito1984sulfur,smrekar2010volcano,shalygin2015volcano} simulation studies \cite{gulcher2020corona}, and the extrapolation of eruptive data from Earth \cite{byrne2022estimate} hypothesize the existence of ongoing, present-day volcanic activity on Venus. Estimates of surface seismicity based on heat flow place rates of Venus seismicity between Earth's and Mars's \cite{lognonne2015quake}. The VERITAS and EnVision missions will investigate surface topography at much higher resolution and examine surface emissivity in several spectral bands \cite{helbert2018venus}, revealing the state of current geologic activity in significantly higher detail. However, no direct measurements of an ongoing volcanic eruption or a venusquake are available, largely because the surface is difficult to observe from orbit to due to a dense shroud of sulfuric acid clouds from 47-70 km altitude \cite{hansen1971clouds} and to inhospitable conditions on the surface, with temperature greater than 460$^\circ$ C and pressure greater than 90 atmospheres \cite{wood1968Venus} that limit the lifetime of landed instruments.

The atmospheric temperature drops steadily with \revI{altitude from the surface to 90-100 km}, and pressure and temperature conditions are much more temperate, and nearly Earth-like in the 50-65 km altitude band \cite{linkin1986thermal}.
This region is within the Venus sulfuric acid cloud layer. While acid survivability is still a concern, thermal conditions are much more friendly for missions inhabiting this region of the Venus atmosphere compared to those that land on the surface.
This is demonstrated in part by the fact that the Soviet Union's VeGa balloons lasted for nearly 48 hours before its batteries were exhausted, which is over 20 times longer than the lifetime of longest-surviving lander, Venera 13, which operated for 127 minutes on the surface \cite{sagdeev1986vega}.
Venus is also a slow, retrograde rotator (with a solar day equivalent to approximately 117 Earth days) and exhibits atmospheric super-rotation, i.e., the atmosphere travels around the planet faster than the planet rotates about its own axis \cite{lebonnois2010superrotation}. The VeGa balloons traversed approximately halfway around the planet over a period of nearly 48 hours. Thus, aerial vehicles in the Venus cloud layer offer the unique possibility of global coverage from an elevated, yet in-situ observation platform.

The next generation of Venus aerial vehicles, \revI{or \emph{aerobots}}, currently under development, will be capable of controlling their altitude and navigating different layers of the Venus atmosphere \cite{hall2021aerobot}. Critically, the ability to change altitude can allow the balloons to navigate to specific sites of interest on the surface, riding stratified winds flowing in different directions \cite{cutts2018}. The ability to exploit buoyancy control and knowledge of wind patterns for accurate guidance of aerial platforms has been effectively demonstrated on Earth \cite{bellemare2020autonomous} for telecommunication applications, and similar techniques have been demonstrated for control of vertically profiling floats in ocean currents \cite{Troesch2018-hm, Dahl2011-lb}. However, crucially, existing approaches rely on accurate knowledge of atmospheric circulation informed by dense, frequent, in-situ and remote measurements - data which is not available for Venus. The work in \cite{Wolf2010Venus} proposes an approach to navigation of Venus aerobots using a stochastic model of the flow field, but it employs an extremely simplified probabilistic model to capture wind variability. \revII{The approach proposed in this paper} extends \cite{Wolf2010Venus} by (i) proposing a rigorous and systematic way of capturing the wind flow probabilistic distribution from time-varying model data, and (ii) extending the concept of operations to multi-balloon systems.

The ability to navigate winds to target specific locations on the surface enables the performance of high-impact in-situ science for the investigation of ongoing volcanic eruptions. For instance, a balloon can take up-close images in many spectral bands on several revisits to a volcanic eruption site, tracking the appearance and evolution of the site. The balloon may also be able to drop an instrumented dropsonde \cite{izraelevitz2020dropsonde} into the eruption plume to sample plume composition at various stages of interaction with the background atmosphere. %

The autonomy of balloon platforms \revII{holds promise to play a significant role in enhancing science returns of future Venus missions.}
Many events of interest may be short-lived and require decisions to be made on the platform itself, rather than waiting for a guidance command with humans in the loop. In this paper, we focus on the study of volcanism on Venus -- that is, success in our approach is measured by the ability of a balloon network to navigate within 50 km horizontal distance of a detected volcanic eruption to perform high-value science.
\revII{
To date, no volcanic eruptions have been directly observed on Venus. Up-close observations of even a single ongoing, or recently concluded, volcanic eruption would provide unprecedented information about the nature of Venus volcanism and the contents of its subsurface. In-situ sampling of the plume of an ongoing eruption, or imaging of the site of an eruption as it occurs or soon after it concludes, would be immensely valuable for understanding Venus' interior. In addition, no two volcanoes are alike, and no two eruptions from the same volcano are alike. Observations of additional eruptions would allow for sampling a diverse nature of volcanic settings and styles -- for instance, varying levels of explosivity, which could point to variations in magmatic volatile content.
The comparison of such eruptions would provide key evidence regarding the distribution of the degree of explosivity of the eruptions, the geographic variation of sub-surface volatile content, and the broader geological and tectonic setting for each volcano.
}

\subsection{Contribution}
Our contribution in this work is threefold.
First, we propose algorithms for \revII{long-range} guidance of an individual altitude-controlled balloon in an uncertain wind field towards regions of interest.
Second, we propose a multi-agent autonomy architecture that integrates autonomous detection of \revI{explosive} volcanic events of interest, information-sharing between multiple balloons and an orbiter, and the proposed guidance algorithms, with the goal of enabling close-up observations of volcanic events. 
Finally, we assess the performance of the proposed architecture, and its sensitivity to fleet size, detector performance, and event frequency, through high-fidelity simulations in a variety of scenarios.
Our results show that the approach can result in an increase of up to 63\% in the number of follow-on observations compared to passive drifters, and a 16\% increase compared to ground-in-the-loop guidance; the approach is robust to uncertainty in the wind field, to large changes in the frequency of volcanic events and in the sensitivity of the microbarometer detectors, and to different numbers of aerial platforms.
\revII{We stress that the goal of this work is not to capture the system-level impact of on-board autonomy in terms of cost, complexity, or risk. Rather, we focus on identifying the scientific potential of autonomous guidance; we envision that this information will be critical to guide follow-on system engineering studies that will explore the complex design space of future Venus explorers.}

\subsection{Organization}
The rest of this paper is organized as follows. In Section \ref{sec:problem}, we formally define the problem of multi-agent balloon navigation for monitoring of volcanic eruptions. In Section \ref{sec:approach}, we present the proposed approach to guidance of teams of buoyancy-controlled balloons in a partially-unknown wind field. The approach is evaluated through detailed numerical simulations: Section \ref{sec:experiments:setup} presents the simulation setup, and Section \ref{sec:experiments:results} reports the results. Finally, in Section \ref{sec:conclusions}, we present our conclusions and identify directions of interest for future research.

\begin{figure}[h]
    \centering
    \includegraphics[scale=0.24]{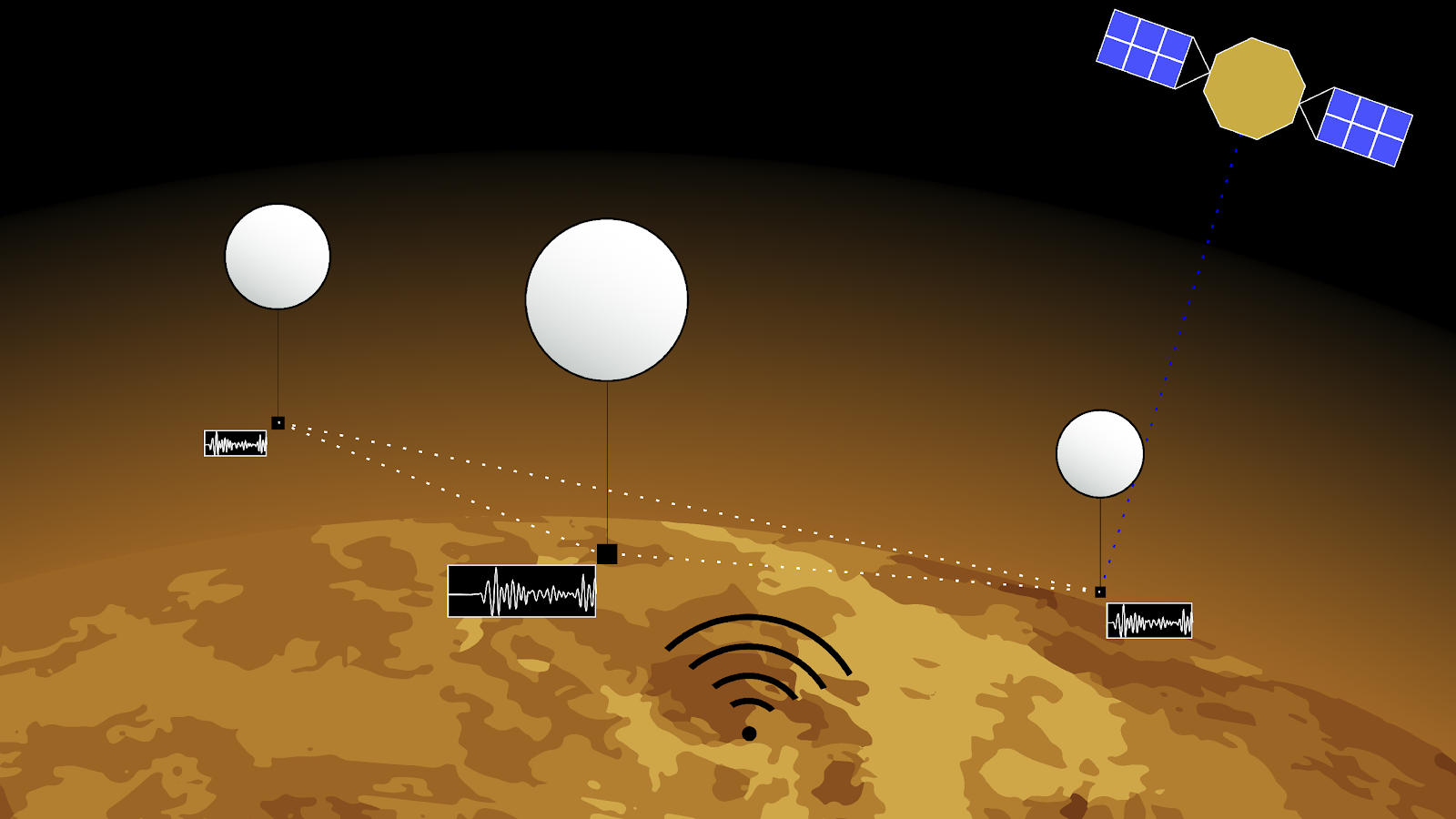}
    \caption{Concept of Operations: a conceptual mission with three balloons aided by one orbiter, which acts as a communication relay between the balloons, supports detection of \revI{explosive} volcanic events, and relays commands from Earth.}
    \label{fig:conops}
\end{figure}

\section{Problem Statement, Concept of Operations, \revI{and Assumptions}}
\label{sec:problem}

In this section, we present in detail a concept of operations for a notional multi-balloon mission to Venus aimed at identifying and studying volcanic activity (shown in Figure \ref{fig:conops}).

We consider a notional mission consisting of three variable-altitude, buoyancy-controlled balloons aided by an orbiter.

\revI{The high-level design motivating this concept of operations is inspired by the JPL Venus aerobot concept \cite{cutts2022exploring,izraelevitz2022terrestrial}, a buoyancy-controlled balloon 12-15m in diameter capable of carrying 100-200 kg payloads in Venus's atmosphere at altitudes between 52-62 km.}

We assume that eruptions are detected and geolocated by the balloons \revI{by measuring} atmospheric pressure fluctuations generated by direct injection of acoustic energy into the atmosphere, or through seismo-acoustic coupling of ground motion excited by an eruption. Volcanic eruptions on Earth and ground motion from seismic activity have been detected and characterized using linear arrays of balloon-based barometers \revI{installed on a single balloon} \revI{\cite{matoza2017,krishnamoorthy2019,brissaud2021infrasound,garcia2022infrasound}}. New techniques that monitor the balloon's dynamic response to an incident pressure perturbation to determine its direction are also under development \revI{\cite{garcia2021glanes,bowman2022aero}}. In this work, we assume that direction-finding techniques are mature enough that instruments on board individual balloons are able to recognize and independently estimate the location of an eruption when they detect one. The likelihood of detecting an event varies depending on the distance between the balloon and the event, and on the event's duration. Volcanic events occur according to a spatial and temporal distribution that is \emph{not} assumed to be known a priori; their duration is also not known a priori. Events can only be detected while they are active. 

An orbiter acts as a communication relay platform. In addition, the orbiter is assumed to host optical instruments that can detect and locate explosive volcanic events by observing fluctuations in the 1.27 $\mu$m airglow layer and the 4.32 $\mu$m dayside CO\textsubscript{2} non-local-thermal-equilibrium (non-LTE) emissions, as proposed in the VAMOS \revI{(Venus Airglow Measurements and Orbiter for Seismicity)} mission concept \cite{didion2018}. \revI{We note that there exist techniques that may also be able to detect non-explosive volcanism from orbit using emissivity mapping or repeat-pass synthetic aperture radar (SAR) interferometry \cite{smrekar2022veritas}.} Balloons can also directly communicate with each other via radio-frequency (RF) links whenever they are within line-of-sight of each other. 

Once a volcanic event is identified and localized, the balloons are tasked with visiting the event from up close in order to perform high-value, in-situ science (e.g., deliver a dropsonde into the volcanic plume,  sample the plume, or capture multispectral imagery). To do this, the balloons control their altitude by changing their buoyancy: when a balloon reaches equilibrium with the surrounding atmosphere in terms of buoyancy, it floats in the direction \revII{of} the wind. Since the wind direction varies with altitude, changing a balloon's buoyancy, hence its altitude, can be used to control the balloon's direction of motion. \revI{In this work, we assume that the balloons can safely control their altitude between 47 km and 63 km above Venus's surface, broadly in line with the design of JPL's aerobot mission concept \cite{cutts2022exploring,izraelevitz2022terrestrial},  with a controlled ascent and descent rate of 1 km/h.}

\revII{We focus on long-range guidance of balloons to volcanic events of interest, and we consider an event to be visited when a balloon is within a 50 km horizontal distance of the event. We envision that, at this range, existing techniques for fine guidance and station-keeping (e.g., \cite{bellemare2020autonomous}) could be adapted for terminal guidance to the event. Explicit modeling of this terminal approach phase is left as future research. 

In order to assess the performance of the proposed approach, we measure both the total number of event visits performed by the balloons (i.e., the number of occurrences when a balloon comes within 50 km of an event of interest), and the number of distinct events visited. We consider a time span of 60 days, corresponding to the proposed lifetime of the Aerobot mission concept \cite{cutts2022exploring,izraelevitz2022terrestrial}. }

A number of numerical models are available to simulate atmospheric circulation on Venus (e.g., \cite{lebonnois2010superrotation,scarica2019validation}). However, these models carry considerable uncertainty due to insufficient characterization of complex processes within Venus' atmosphere, primarily due to a lack of in-situ data. 
Accordingly, we do \emph{not} assume that the wind velocity profile is known. Rather, we assume that a \emph{probabilistic model} is available that characterizes the \emph{distribution} of likely wind velocities that a balloon will encounter at a given location within Venus's atmosphere. (We remark that this formulation is highly general: as an extreme example, lack of any knowledge about atmospheric circulation in a given region of the atmosphere can be captured through \revI{a uniform distribution for the wind's direction, and a Jeffreys prior \cite{jeffreys1946invariant} for wind speed.}) %

In this paper, we do not explicitly model the process of performing high-value proximity science to ensure generality; rather, we associate proximity with the event with a positive reward. The high-value science can be performed both if the event is active, and if the event has concluded since, even if a volcano is no longer active, it is of scientific interest to observe the geologically young features following a recent eruption that have not been weathered.

\section{Autonomous Guidance of Teams of Balloons in an Uncertain Wind Field}
\label{sec:approach}

\begin{figure}[h]
    \centering
    \includegraphics[width=\textwidth]{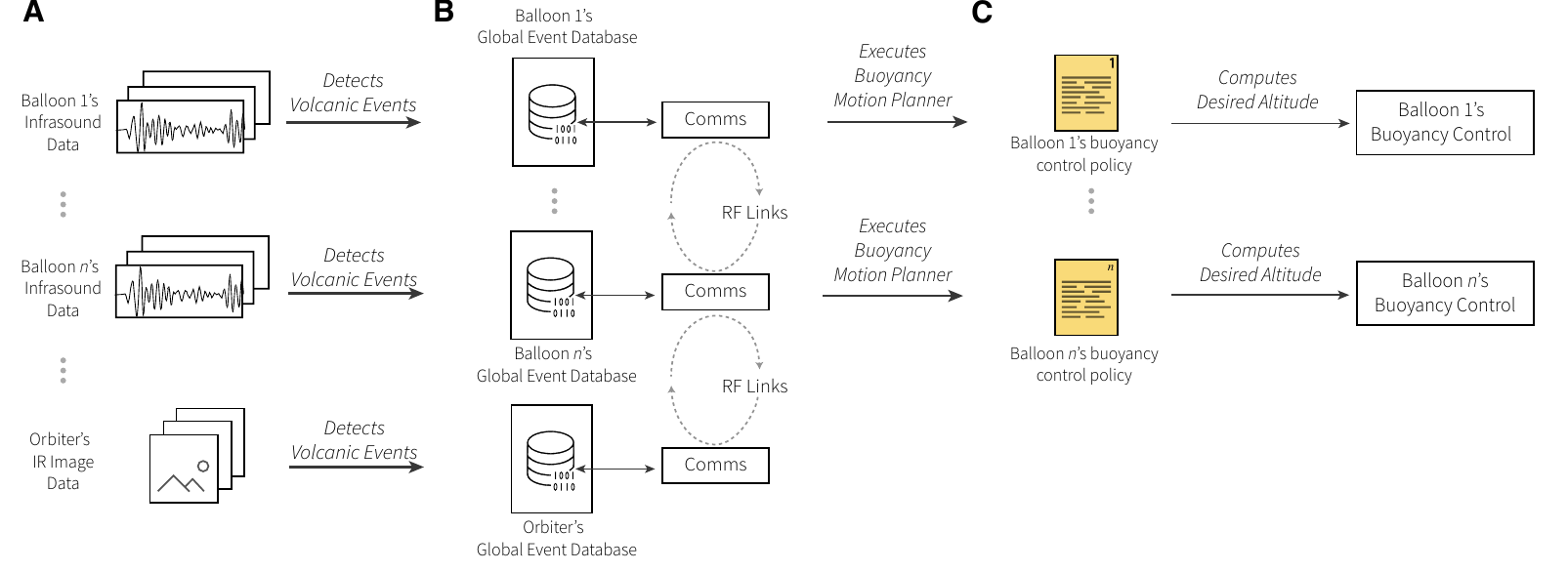}
    \caption{Proposed autonomy architecture. Balloons detect volcanic events by measuring atmospheric pressure fluctuations with on-board barometers; the orbiter also detects \revI{explosive} volcanic events by observing fluctuations in the airglow layer and CO\textsubscript{2} emissions. Events are stored in a global event database, which is synchronized among the agents over communication links. Balloons then compute an optimal buoyancy control policy to maximize the likelihood of visiting events of interest, based on a stochastic model of atmospheric circulation.}
    \label{fig:approach:architecture}
\end{figure}

We propose a multi-agent software architecture designed to realize this concept of operations. The architecture, shown in Figure \ref{fig:approach:architecture}, consists of three parts: an eruption detection module based on infrasound remote sensing, a shared event database that keeps track of detected eruptions, and a buoyancy motion planner. Each element of the architecture is executed on each balloon, and also on the orbiter; the event database is synchronized between the agents whenever a communication link is available.

\subsection{Eruption Detection Model}
\label{sec:approach:volcano:detection} 

We assume that a balloon is able to detect and localize a volcanic eruption whenever the balloon is within a given range of the eruption, which we denote as the ``detection radius''. We model the detection radius as a function of the eruption's volcanic explosivity index (VEI, \cite{newhall1982vei}), a qualitative measure of the eruptive nature of a volcanic eruption that is estimated based on the intensity of the release of energy and matter from a volcanic eruption.
The proposed estimate is based on the work in Dabrowa et al. \cite{dabrowa2011plume}, which studied the potential for long-range infrasonic detection of volcanoes on Earth and generated an empirical relationship between the plume height and detection distance of a volcanic eruption on Earth.
The VEI determination for any volcanic eruption incorporates plume height as a factor. We combined the relationship between plume heights and VEI with the relationship between detection radii and plume heights to generate the values for the detection radius in Table \ref{tab:model:detection-radius}. Since the range of plume height associated with a given VEI can be fairly large (e~.g~, VEI 4 eruptions typically have plume heights between 10 and 25 km), we picked the upper limit of plume height for each VEI as the value to use for determining detection radius.\\
Our approach \revI{inherently possesses large uncertainties} -- there is significant scatter in data from \cite{dabrowa2011plume} resulting from atmospheric variability. \revI{While balloon-based acoustic detection of artificially and naturally generated seismic signals has been demonstrated on Earth \cite{brissaud2021infrasound,bowman2021explosion}, the intensity of volcanic phenomena on Venus, and the acoustic coupling between such phenomena and the atmosphere, are less understood. The study in \cite{dabrowa2011plume} also focused on the detection of eruptions from ground stations, which can often result from acoustic signals following very complex propagation paths, including several reflections. Additionally, plume heights associated with any eruption are extremely variable, thus, the plume height ranges specified by \cite{newhall1982vei} carry significant uncertainty.} Further, plume dynamics are also expected to be very different on Venus compared to Earth, as a result of the high-pressure atmosphere.  \\
\revI{The type of volcanism thought to be present on Venus also merits some discussion. The 90-bar atmosphere near the surface may act as strong barrier to explosive volcanism and allow only effusive volcanism, reducing the atmospheric signature. However, the volatile content in the Venusian mantle is also unknown and may still cause explosive volcanism (a Krakatoa-style eruption was hypothesized by \cite{esposito1984sulfur}, for instance), but  at a reduced rate compared to Earth. Indeed, based on the morphology and radar backscatter of volcanic edifices and surrounding lava flows, Ganesh et al. \cite{ganesh2021effusive} have hypothesized that effusive volcanism is the dominant mode on Venus. It must also be noted though, that volcanic processes trigger co-eruptive seismic signals, which are coupled into the atmosphere 60 times stronger on Venus compared to Earth \cite{garcia2005venus}, resulting in much larger acoustic signals from surface activity of the same intensity. In addition, balloons are much better positioned to capture acoustic signals trapped in atmospheric ducts, by virtue of moving within the same ducts, compared to ground stations.} \\

Owing to these uncertainties, we did not scale detection radii estimates from Earth to Venus. However, in recognition of the heuristic nature of the estimates in Table \ref{tab:model:detection-radius}, we assessed the sensitivity of the proposed approach to large changes in the detection radius and in the frequency of volcanic events in Sections \ref{sec:results:detection_radius} and \ref{sec:experiments:results:event_frequency} respectively.
\\
\revI{For the remainder of this manuscript, the term ``volcanic events" implies ``explosive" volcanic events, as the statistics for occurrence and detection radii are primarily derived from Earth-based events. However, we do so with the caveat the effusive volcanism may also be detected using acoustic methods through the mechanisms listed above.
We recognize that VEI is an all-encompassing, qualitative metric of the strength of a volcanic eruption. Additionally, in this work, we do not consider several alternative, and highly promising, ways of detecting volcanic eruptions on Venus from airborne platforms based on emission of heat or volatiles. 
These choices are motivated by the focus of the paper on the design of autonomous policies to perform follow-up observations {\emph{after}} detection. Considering higher-fidelity models for detections or events of interest, or non-acoustic methods for airborne detection of events of interest, are interesting directions for future work. 

Lastly, we also assume that the technology to autonomously geolocate the site of the eruption from the acoustic signal is mature, such that each balloon is independently able to find out where the eruption occurred by analyzing the detected infrasound signal on board.}

\revI{
\begin{table}[h]
\centering
\begin{tabular}{c|p{2.3cm}p{2.3cm}p{2.6cm}|p{2.3cm}}
VEI & Plume height [km] & Ejecta [m$^3$] & Total acoustic energy [MJ] & Detection radius [km] \\
0 & $\leq 0.1 $  & $\leq 10^4$            &    $200$                  & 110 \\
1 & 0.1-1           & $10^4-10^6$         &   $1.8 \cdot 10^4$   & 370 \\
2 & 1-5              & $10^6-10^7$         &   $4.3 \cdot 10^5$   & 1400 \\
3 & 3-15            & $10^7-10^8$         &   $3.7 \cdot 10^6$   & 4100 \\
4 & 10-25          & $10^8-10^9$         &   $10^7$                  & 6800 \\
5 & $20-35$      & $10^9-10^{10}$     &   $2\cdot 10^7$       & 9500 \\
6 & $\geq 30$   & $10^{10}-10^{11}$ &   $2.5\cdot 10^7$    &  11000
\end{tabular}
\caption{Detection radius as a function of an eruption's VEI. \revI{The relationship between VEI, volume of ejecta, and plume height is derived from \cite{newhall1982vei}; the relationship between plume height and detection radius is based on the estimate presented in \cite{dabrowa2011plume}. Both estimates are for Earth volcanoes. Sensitivity of the proposed approach to variations in detection radii are explored in Section \ref{sec:results:detection_radius}. }}
\label{tab:model:detection-radius}
\end{table}
}

\revI{We also assume that the orbiter is equipped with a wide-angle optical instrument
able to detect eruptions of VEI $\geq 2$ whenever they are visible to the orbiter; detection is performed through observations in the 1.27 $\mu$m airglow layer and the 4.32 $\mu$m dayside CO\textsubscript{2} non-local-thermal-equilibrium emissions, as proposed in the VAMOS mission concept.}

\subsection{Single-balloon buoyancy motion planner} %
\label{sec:approach:mdp}

The motion planner guides aerial platforms towards events of interest by controlling the balloon's buoyancy so as to steer it towards altitudes that are most likely to present favorable winds.

We pose the motion planning problem for the balloon as a continuous-state Markov Decision Process (MDP) \cite{bertsekas2017dynamicI}.

Intuitively, a Markov Decision Process allows to compute an optimal \emph{policy} that provides the best action that an agent should take for each state that the agent may assume. In this paper, the agent is the balloon; the agent's states are the set of longitudes, latitudes, and altitudes that the balloon may occupy; and the actions capture the decision to climb, descend, or maintain a constant altitude. The optimal policy is computed so as to maximize the integral of a reward function along the balloon's trajectory (discretized according to a discrete time step $\delta$). In this work, the balloons receive a positive reward for visiting detected volcanic events, and a much smaller negative reward for changing altitude, which models the energy cost required to change the balloon's buoyancy. The policy is computed based on a \emph{stochastic} model of the system's transitions, which capture the likelihood that a balloon will drift to a specified latitude, longitude, and altitude from a given starting position, given the autonomy's decision to climb, descend, or maintain altitude.  The proposed approach is closely related to the one described in \cite{RossiBranchEtAl2021} for control of under-ice drifters.

 To formalize the MDP, we describe its states, actions, transitions, rewards, and final states. 
 
\subsubsection{States} The state of the balloon is represented by its location in Venus's atmosphere, namely, its longitude \revI{$\varphi$}, latitude \revI{$\lambda$}, and altitude \revI{$h$}. The balloon's altitude \revI{$h$} is constrained to lie between prescribed minimum and maximum altitudes $\underline h$ and $\overline h$, corresponding to the region of the Venusian atmosphere where the balloon can safely operate. Formally, the state of the Markov decision process is  $\mathcal{S} = \{ (\varphi, \lambda, h): \varphi \in [-\pi, \pi], \lambda\in [-\pi/2, \pi/2], h\in [\underline h, \overline h]\}$.

\subsubsection{Actions} The goal of the decision process is to select the altitude that the balloon should move to at a given location. At every time step, we allow the balloon to take one of three possible actions: climb by \revI{1000} meters, descend by \revI{1000} meters, or maintain the current altitude. The new altitude is constrained to lie between the minimum and maximum allowed altitudes. Rigorously, the action set for a given state $(\varphi, \lambda, h)$ is $\mathcal{A}(\varphi, \lambda, h) = \{h, \min(h+\revI{1000} m, \overline h), \max(h-\revI{1000} m, \underline h)\}$. %
\revI{We remark that, when the policy is executed, the balloons will change their altitude gradually over the duration of a time step $\delta$, which is set to one hour in the simulations.}

\subsubsection{Transitions}
\label{sec:approach:mdp:transitions}
 The balloon is modeled as a semi-Lagrangian tracer:  the balloon moves along with the wind in the horizontal plane, and the vertical dynamics are assumed to be fully controllable by the buoyancy control. We assume that a \emph{stochastic} model is available that describes the likely distribution of wind speeds at every point in the domain of interest. Rigorously, the wind velocity at a given location $v(\varphi, \lambda, h)$ is a random variable characterized by a probability distribution $P( v(\varphi, \lambda, h)): \mathbb{R}^3\mapsto [0,1]$.

We estimate the probability distribution of the wind velocity by fitting a numerical model of Venus atmospheric circulation. 
Numerical simulations are generally unable to accurately reproduce atmospheric circulation \emph{point-wise}, i.e., at a specific time and location, due to dependency on unknown initial conditions and on unmodeled forcing terms. Our key assumption is that the empirical temporal distribution of velocities in a given region of the atmosphere produced by a numerical model across its integration horizon is representative of the probability distribution of wind velocities that would be encountered by a vehicle traversing that region; that is, that the numerical simulation is \emph{statistically representative} of the underlying process. Intuitively, if the numerical model predicts highly consistent wind directions and velocities in a given region, the actual probability distribution of flow velocities in the region should match the numerical prediction with high probability; if the numerical models predicts highly varying flow velocities in a given region, the actual flow velocity encountered by a balloon should also admit higher uncertainty.

Rigorously, we assume that a numerical model of the time-varying atmospheric circulation is available that specifies the wind velocity \revI{$v_N$} for all points of interest in the atmosphere \revI{$\varphi,\lambda,h$} \revI{for every time instant $t$ in  a given, representative time horizon $[t_0, t_F]$}: specifically, the numerical model provides a function $\hat v_N(t,\varphi,\lambda,h): \{[t_0, t_F], [-\pi, \pi], [-\pi/2, \pi/2], [\underline h, \overline h]\} \mapsto \mathbb{R}^3 $.
Typically, the numerical model provides flow velocities on a grid of discrete points; the function $\hat v_N$ can then be obtained as an interpolation of the flow velocities on the grid.

To obtain a stochastic model of the wind field, we propose simply using the empirical distribution provided by the numerical model: %

\begin{equation}
P(v(\varphi, \lambda, h) = v)   = \frac{ \int_{t_0}^{t_F} 1_{\hat{v}_N (t,\varphi, \lambda, h)=v} dt}{\int_{\varphi=-\pi}^{\pi} \int_{\lambda=-\pi/2}^{\pi/2} \int_{t_0}^{t_F} 1_{\hat{v}_N (t,\varphi, \lambda, h)=v} dt} %
\label{eq:approach:velocity-probability}
\end{equation}
where $1_{x}$ is a function assuming value 1 if $x$ is true and $0$ otherwise.

The probability of transitioning from state $(\varphi, \lambda, h)$ to state $(\varphi', \lambda', h')$ with  action \revI{ (i.e., commanded altitude) $\hat h$} can then be computed as
\begin{align}
& P((\varphi', \lambda', h')| (\varphi, \lambda, h), \revI{\hat h}) = \nonumber \\
& \quad 1_{\revI{\hat h} = h'} \cdot \int_{\zeta=\underline h}^{\overline{h}} P\left(v(\varphi, \lambda,h) = \left[\frac{(\varphi' - \varphi)(\revI{R_\Venus+}h)\cos(\lambda)}{\delta}, \frac{(\lambda' - \lambda)(\revI{R_\Venus+}h)}{\delta}, \zeta \right]\right) d \zeta, 
\label{eq:mdp:transition_probability}
\end{align}

\revI{where $R_\Venus$ is the radius of Venus}; that is, the probability of reaching state $(\varphi', \lambda', h')$ from state $(\varphi, \lambda,h)$ with commanded altitude \revI{$\hat h$} is equal to the probability of encountering a flow velocity that drags the balloon from $\varphi$ to $\varphi'$ and from $\lambda$ to $\lambda'$ if the commanded altitude equals $h'$, and is zero if the commanded altitude is not $h'$.

\subsubsection{Rewards}
\label{sec:approach:mdp:rewards}

The goal of the motion planner is to drive balloons towards detected volcanic eruptions. To this end, states whose latitude and longitude fall within a 50 km radius of a detected eruption are endowed with a highly positive reward. The location of a detected eruption \revI{$e$} is typically not known perfectly, but rather as a probability distribution \revI{$p_e(\varphi, \lambda): \{  [-\pi, \pi], [-\pi/2, \pi/2]\}\mapsto [0,1]$ that captures the likelihood that eruption $e$ is located at longitude $\varphi$ and latitude $\lambda$}; accordingly, the rewards capture the \emph{probability} that a given state will fall within 50 km of the eruption. In addition, a negative reward \revI{$r_\text{energy}$} is associated with changes of altitude in order to capture the energy cost of changing the balloon's buoyancy.

Formally,
\begin{align}
r((\varphi, \lambda, h), u) = & - |h-u| r_{\text{energy}} + \nonumber \\
&\sum_{e\in \text{eruptions}} \int_{\varphi'=-\pi}^{\pi} \int_{\lambda'=-\pi/2}^{\pi/2} \left (p_e(\varphi', \lambda') 1_{\{d((\varphi, \lambda), (\varphi', \lambda'))\leq 50 \text{km}\}} \right)r_{eruption} \cdot \nonumber \\
& \revI{\left(R_\Venus \cos(\lambda') d\varphi' R_\Venus d\lambda'\right)},
\label{eq:mdp:rewards}
\end{align}
where $d$ denotes the geodesic distance, $r_{\text{eruption}}$ is the reward associated with visiting an eruption, and $r_{\text{energy}}$ is the energy cost associated with a one-meter change in altitude.

\subsubsection{Final States}

Finally, all states with an altitude below the balloon's minimum altitude $ \underline h$ or above the balloon's maximum altitude $ \overline h$ are denoted as final states associated with a large negative reward \revI{$r_\text{altitude}$}, to encourage the balloons to remain within survivable altitude limits.

\subsubsection{Solving the Markov Decision Process}
\label{sec:approach:mdp:adp}

Each balloon solves the proposed Markov Decision Process onboard through approximate dynamic programming \cite{bertsekas2012dynamicII}. The balloon discretizes the state space $(\varphi, \lambda, h)$ in a discrete set of states forming a uniform lattice \revI{$\mathcal{L}$.} We remark that the lattice needs not correspond to the grid used to simulate the atmospheric dynamics. The Bellman equation \revI{for the optimal state value $V^\star$ of} each state in the lattice can be written as
\begin{align}
V^\star(\varphi, \lambda, h) = &\max_{\revI{\hat h}\in\mathcal{A}(\varphi, \lambda, h)} \left( r( (\varphi, \lambda, h), \revI{\hat h}) +\right. \nonumber \\
& \left. \gamma \mathbb{E}_{(\varphi', \lambda', h') )'\sim P((\varphi', \lambda', h')| (\varphi, \lambda, h), \revI{\hat h}) }\!\left[ V^\star((\varphi', \lambda', h')) \right] \right),
\label{eq:optimal-value}
\end{align}
and the optimal commanded altitude $\revI{\hat h^\star}$ for state $(\varphi, \lambda, h)$ can be computed as
\begin{align}
\revI{\hat h}^\star((\varphi, \lambda, h)) = & \arg\max_{\revI{\hat h}\in\mathcal{A}(\varphi, \lambda, h)} \left( r( (\varphi, \lambda, h), \revI{\hat h}) \right. + \nonumber \\
 &\left. \gamma \mathbb{E}_{(\varphi', \lambda', h') )'\sim P((\varphi', \lambda', h')| (\varphi, \lambda, h), \revI{\hat h}) }\!\left[ V^\star((\varphi', \lambda', h')) \right] \right),
\label{eq:optimal-action}
\end{align}

The value of states $(\varphi', \lambda', h')$ not in the lattice is computed by interpolating among states in the lattice, as is standard in approximate dynamic programming. Rigorously, denote as $\tilde{\mathcal{N}}(\varphi', \lambda', h')\subseteq \mathcal{L}$ the set of states that form the vertices of the lattice cell that contains $ (\varphi', \lambda', h')$. Then the optimal value of $(\varphi', \lambda', h')$ is approximated as
\begin{subequations}
\begin{align}
&V^\star(\varphi', \lambda', h') = \sum_{\tilde s \in \tilde{\mathcal{N}}(\varphi', \lambda', h')  } \lambda_{\tilde s} V^\star(\tilde s), \quad \text{ where}
\label{eq:optimal-value-approx} \\
&\lambda_{\tilde s} \propto \frac{1}{\|(\varphi', \lambda', h')-\tilde s\|} \quad \forall \tilde s\in \tilde{\mathcal{N}}(\varphi', \lambda', h') \quad \text{and} \quad
\sum_{\tilde s \in \tilde{\mathcal{N}}} \lambda_{\tilde s} = 1
\label{eq:optimal-value-coeffs}
\end{align}
\label{eq:optimal-value-approx-and-coeffs}
\end{subequations}

Equations \eqref{eq:optimal-value}-\eqref{eq:optimal-value-approx-and-coeffs} are solved via value iteration, resulting in an optimal policy that provides an optimal altitude for each longitude-latitude-altitude tuple in the lattice. For states not in the lattice, we use a nearest-neighbor approach, i.e., we select the altitude corresponding to the closest state in the lattice.

\begin{figure}[h]
\centering
\includegraphics[width=0.32\textwidth]{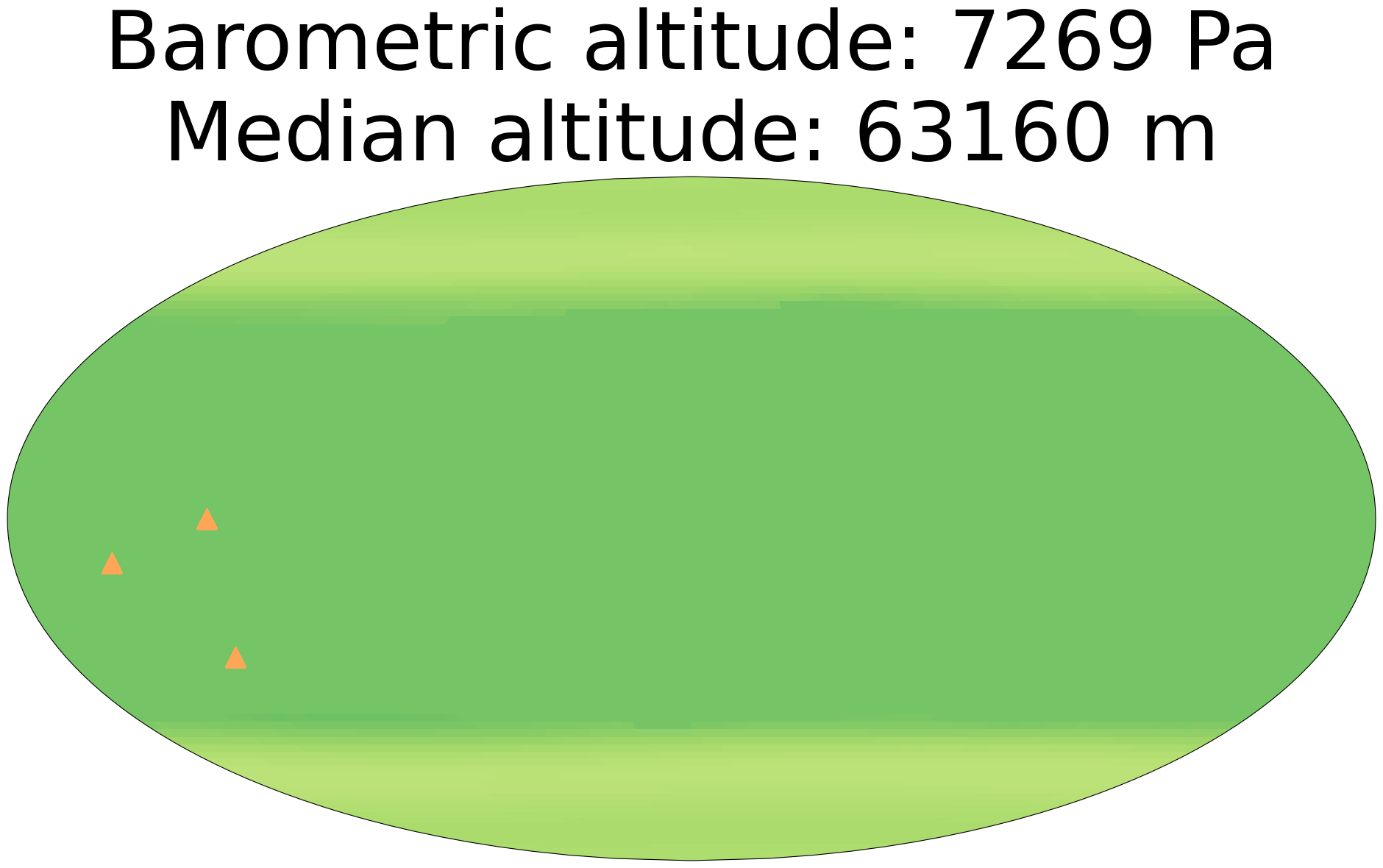}
\includegraphics[width=0.32\textwidth]{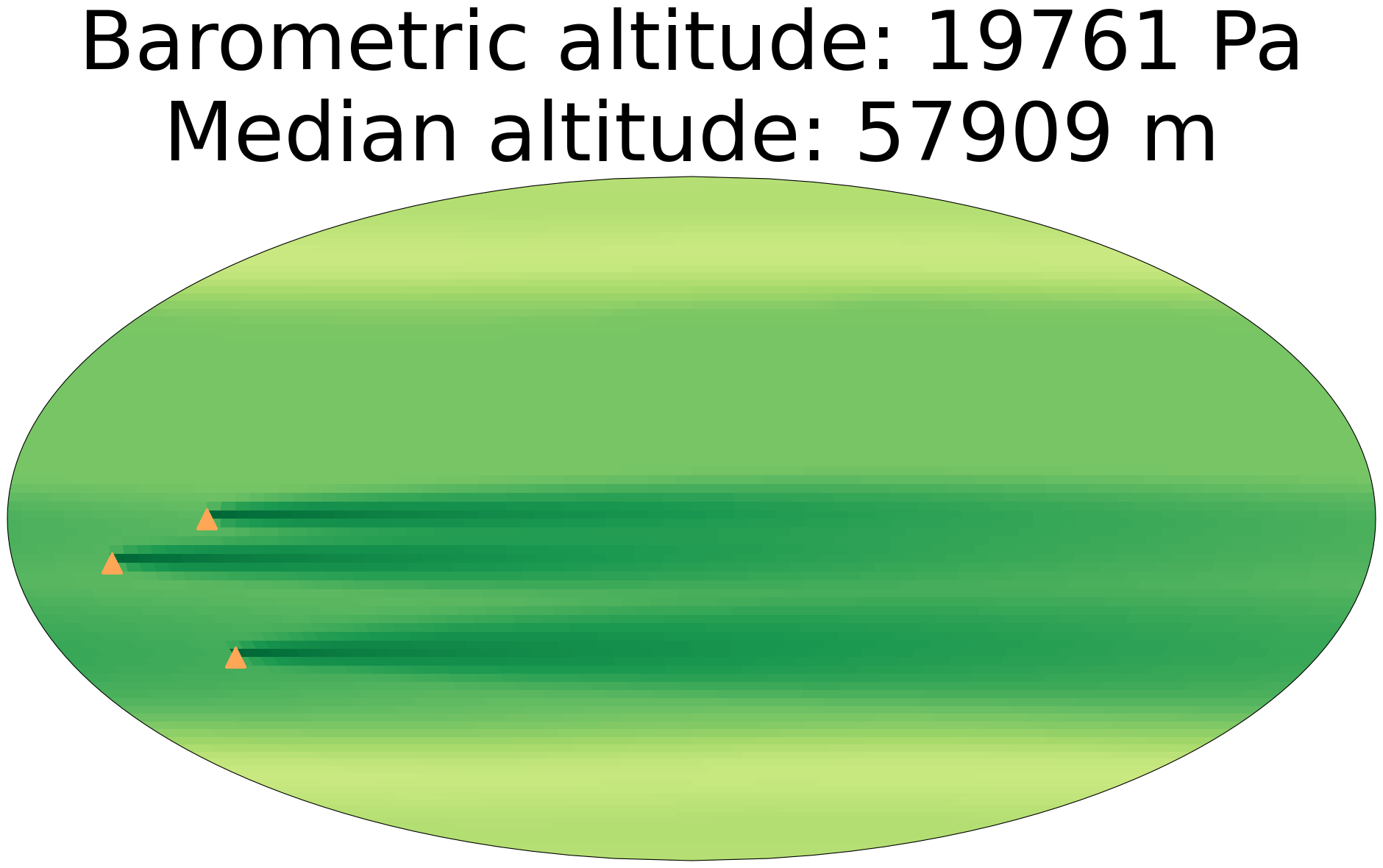}
\includegraphics[width=0.32\textwidth]{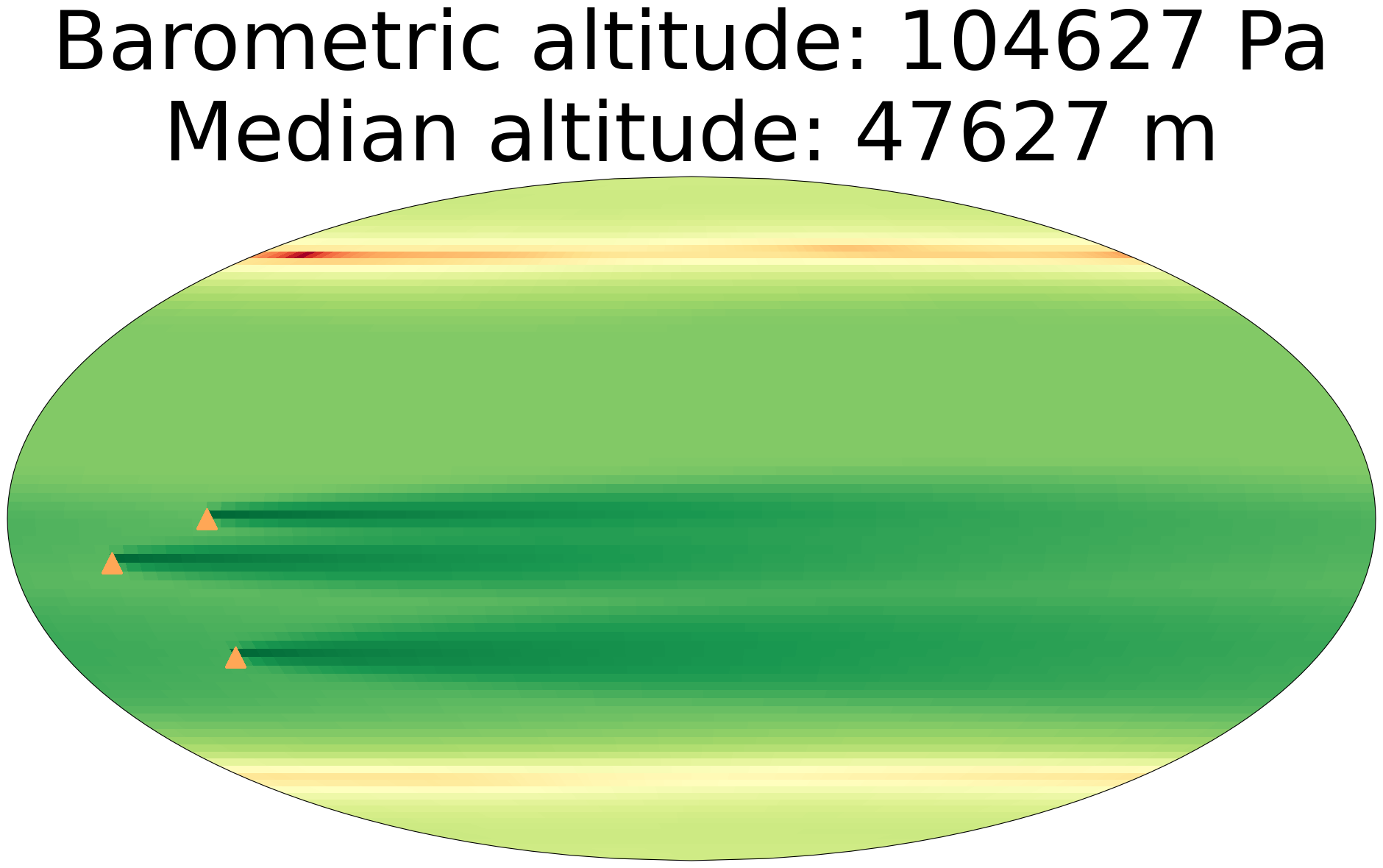}\\
\includegraphics[width=.89\textwidth]{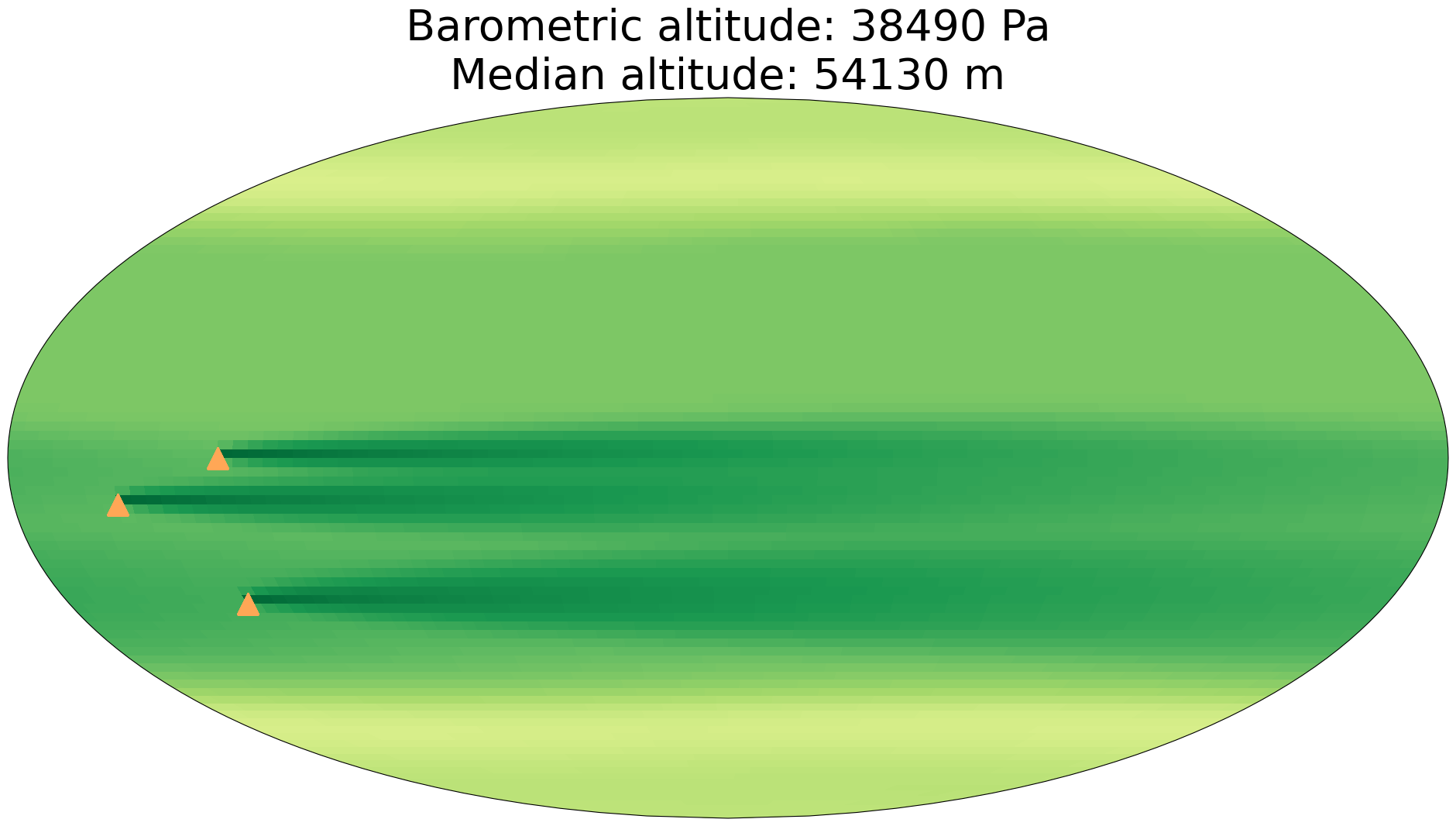}
\includegraphics[width=0.09\textwidth]{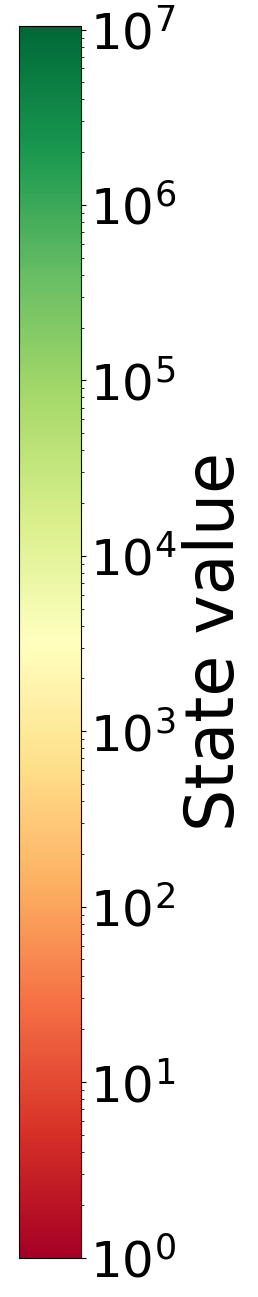}
\caption{State values $V^\star$ for three randomly-selected event locations (denoted by orange triangles) and for selected altitudes.}
\label{fig:mdp:values}
\end{figure}

\begin{figure}[h]
\centering
\includegraphics[width=0.24\textwidth]{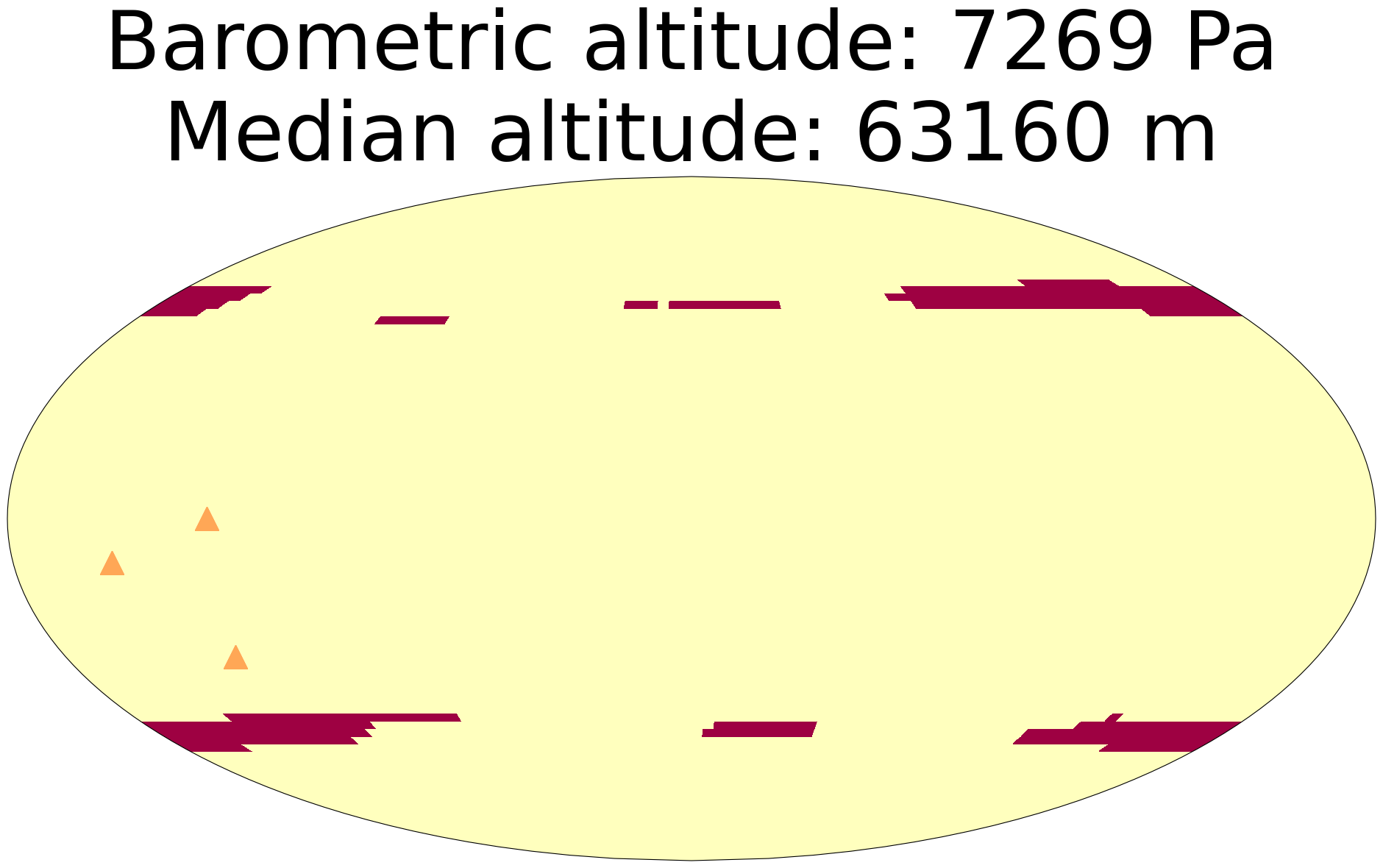}
\includegraphics[width=0.24\textwidth]{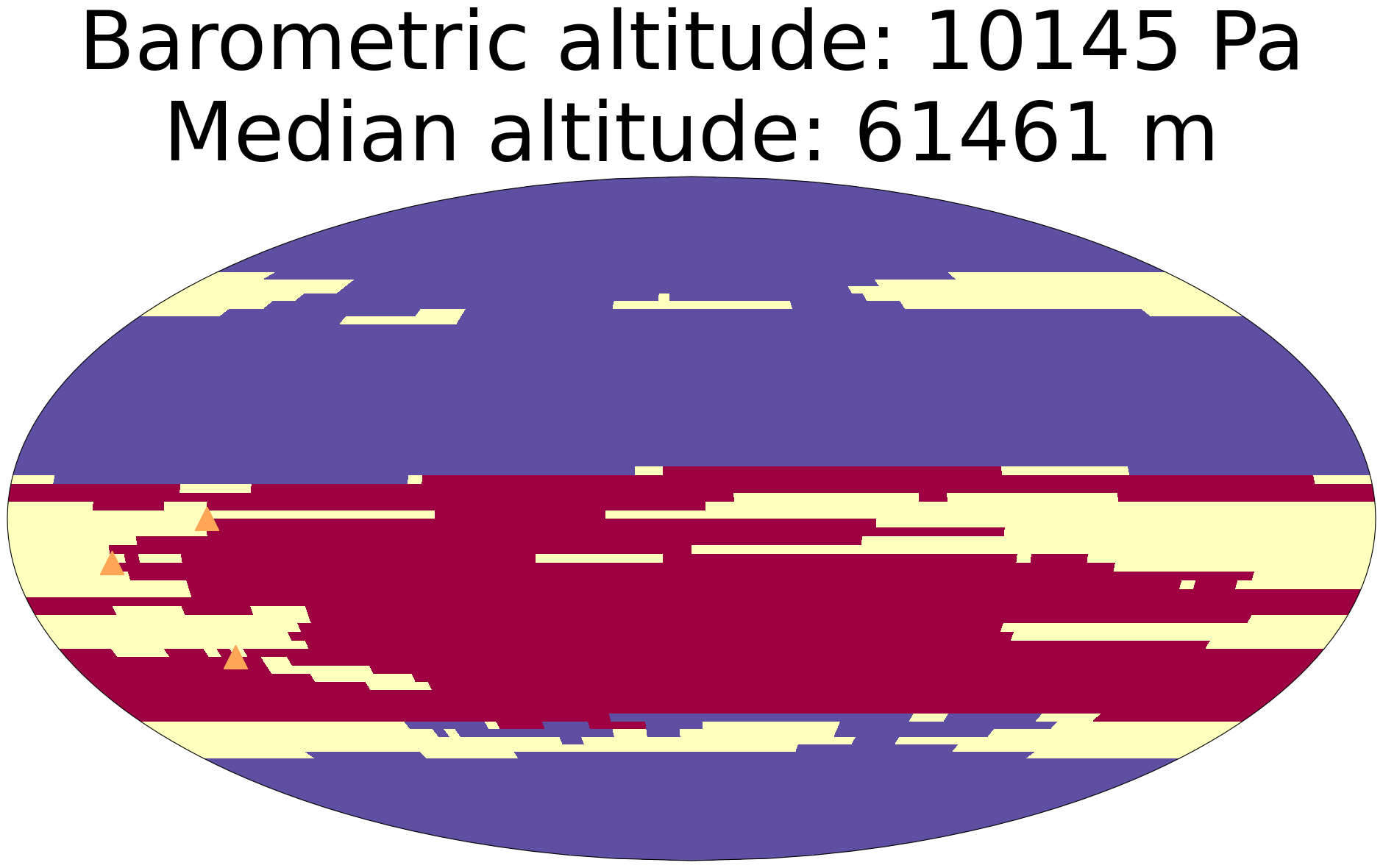}
\includegraphics[width=0.24\textwidth]{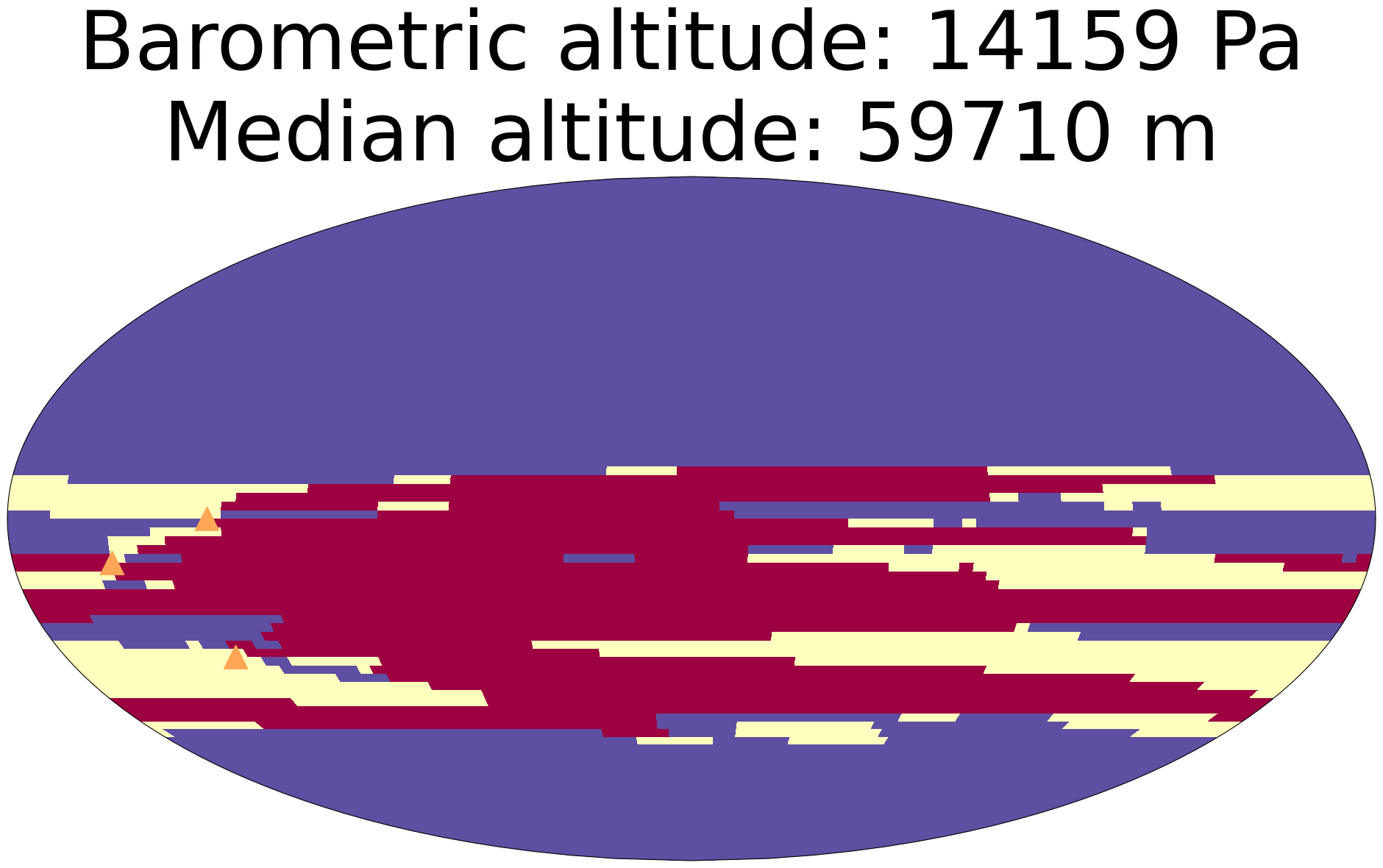}
\includegraphics[width=0.24\textwidth]{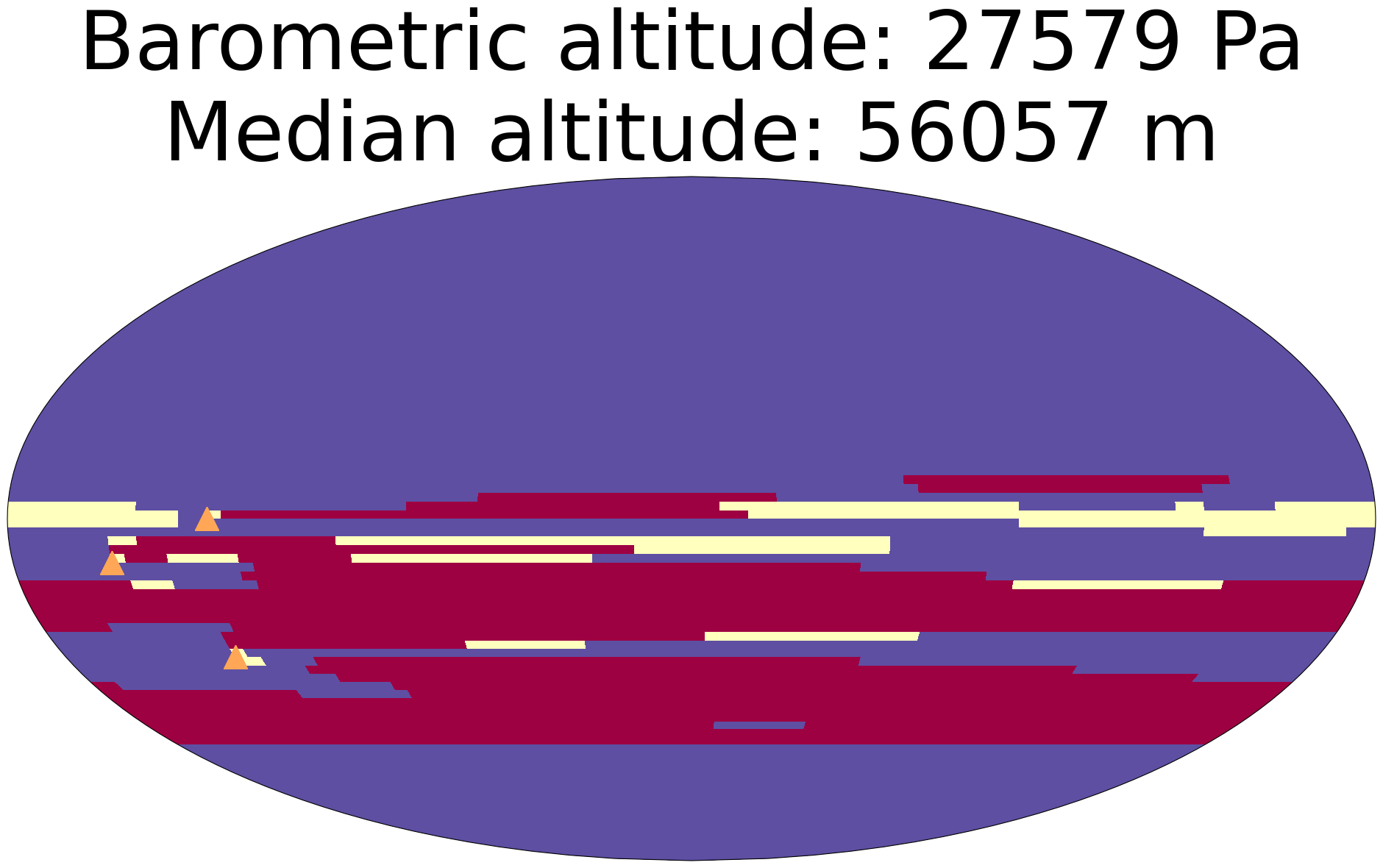}\\
\includegraphics[width=0.24\textwidth]{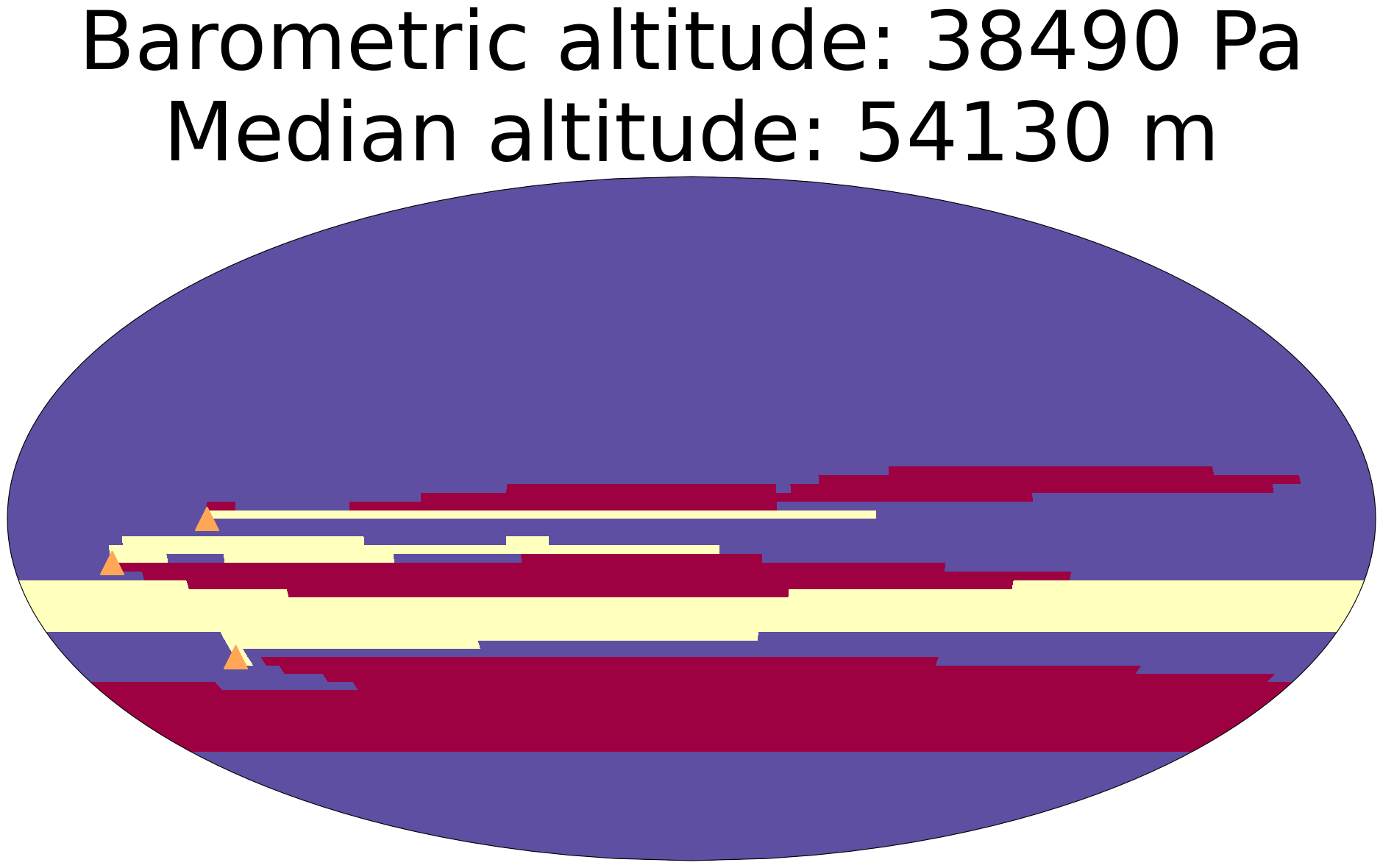}
\includegraphics[width=0.24\textwidth]{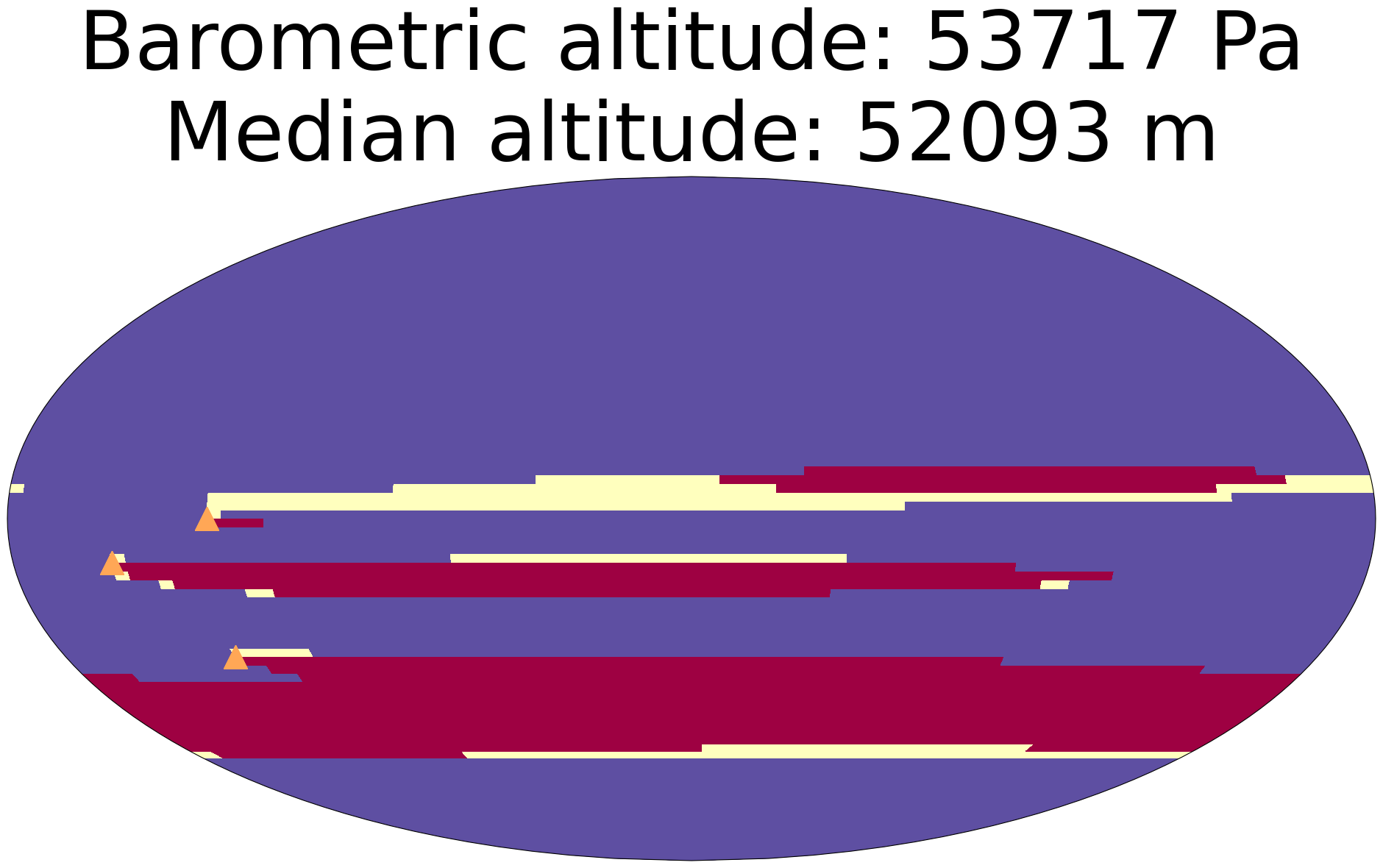}
\includegraphics[width=0.24\textwidth]{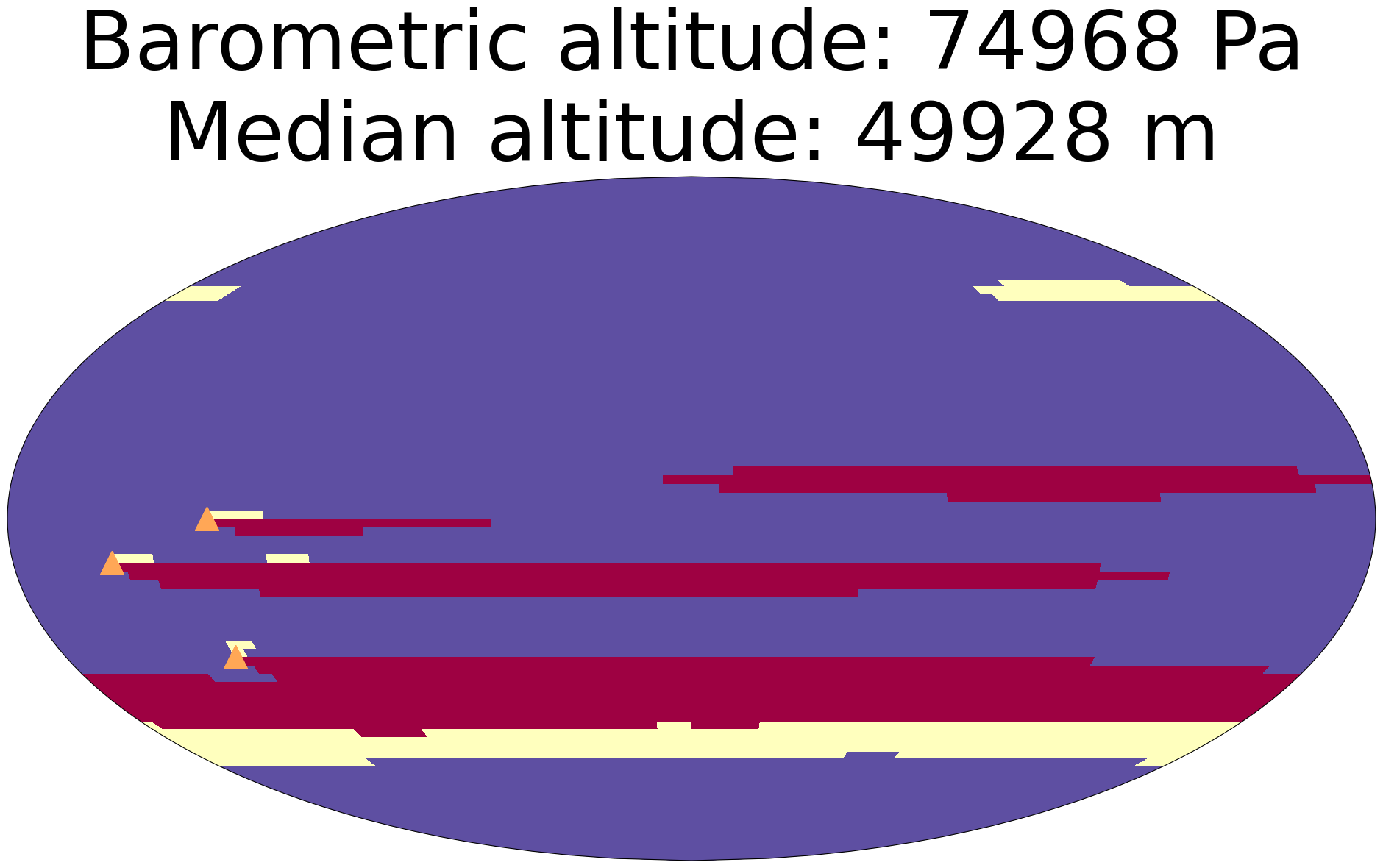}
\includegraphics[width=0.24\textwidth]{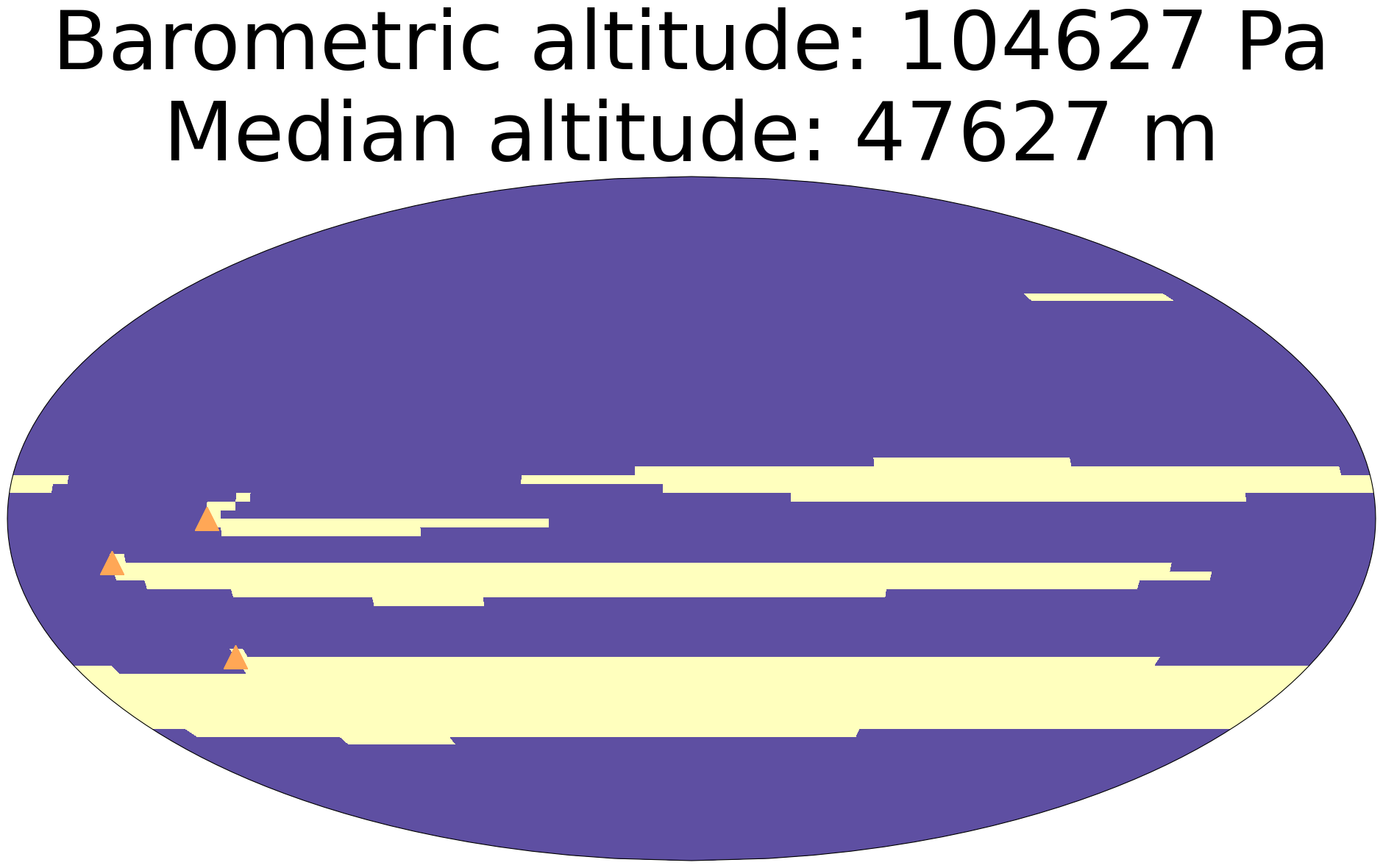}\\
\includegraphics[width=0.86\textwidth]{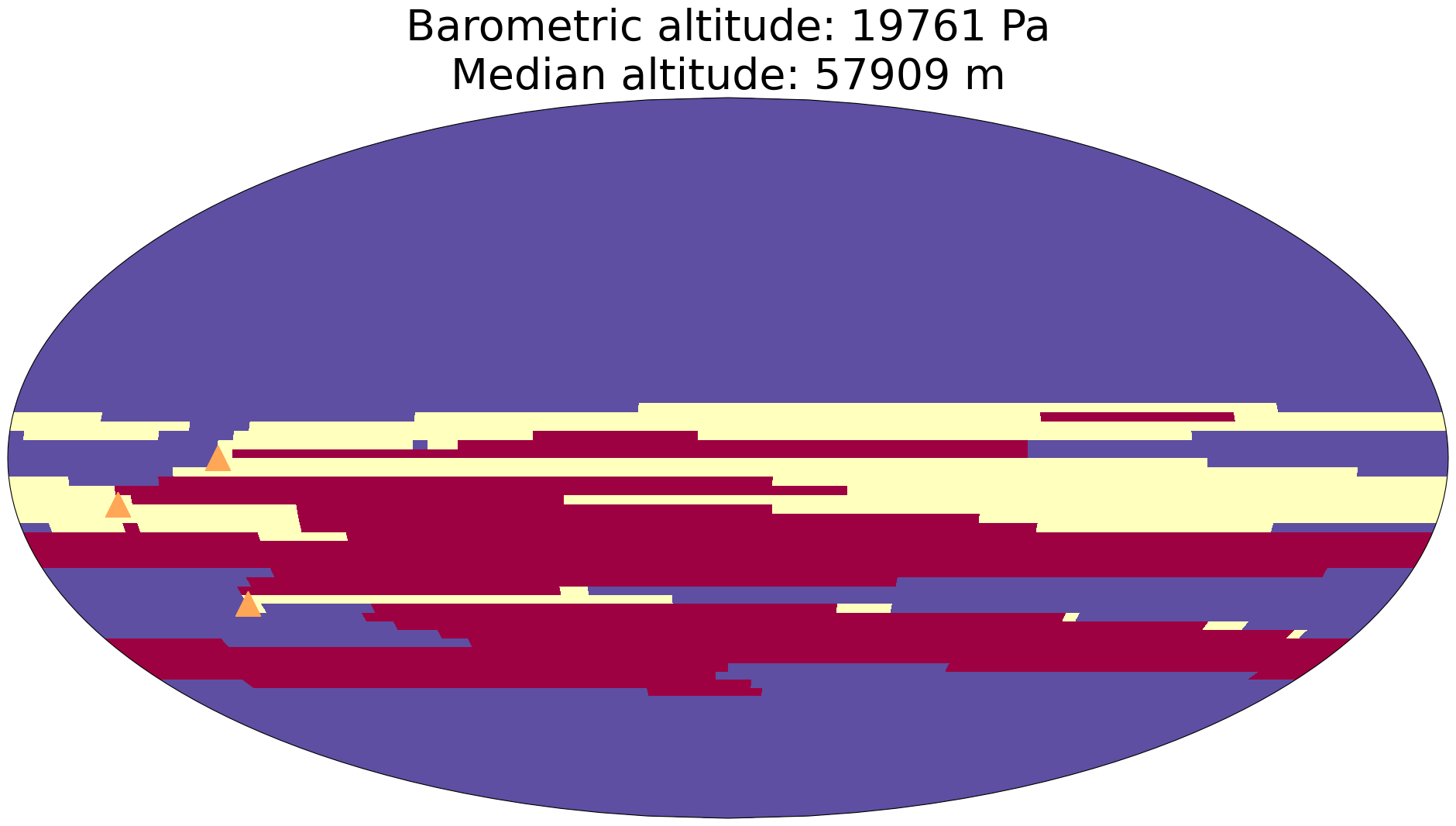}
\includegraphics[width=0.123\textwidth]{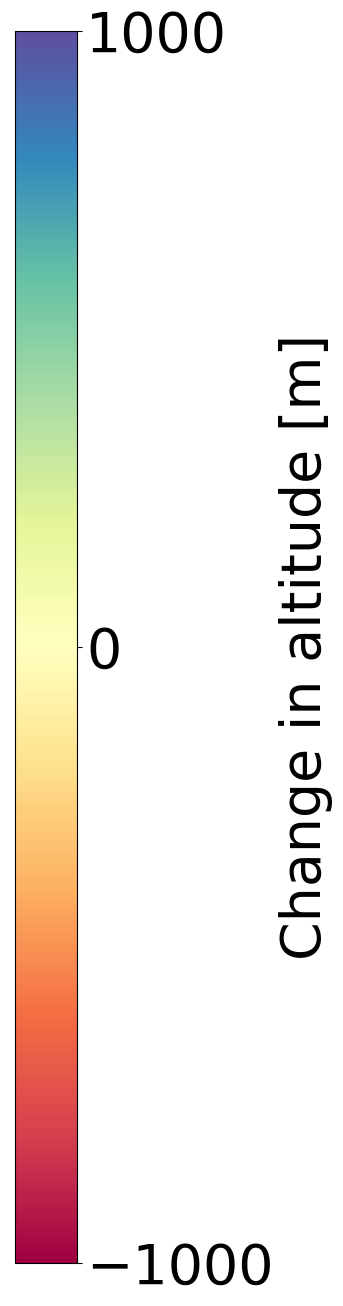}\\
\caption{Optimal policy $\revI{\hat h}^\star$ for three randomly-selected event locations (denoted by orange triangles).}
\label{fig:mdp:policy}
\end{figure}

Figures \ref{fig:mdp:values} and \ref{fig:mdp:policy} shows the optimal state values $V^\star$ (i.e., the expected optimal reward that will be obtained by following the optimal policy from a given state) and the optimal policy $\revI{\hat h}^\star$ for a set of three randomly-selected event locations.

\subsection{Event Database} %

The policy outlined in Section \ref{sec:approach:mdp} can drive a \emph{single} balloon towards eruptions whose location is known. In order to employ the policy in a multi-agent setting, we assume that the balloons share event detections with each other whenever a communication link is available, either through line-of-sight balloon-to-balloon communication, or using the orbiter as a relay.

Specifically, each balloon and the orbiter are endowed with an event database that keeps track of the location of detected eruptions. Whenever two balloons, or a balloon and an orbiter, are able to communicate with each other, they compare their event databases, and each agent updates its database with eruptions detected by the other agent. Since it is of interest to visit past eruptions as well as active ones, events are never removed from the database. 

\subsection{Autonomous Operations } 

The overall architecture and concept of operations is shown in Figure \ref{fig:approach:architecture}. As
the balloons drift through the Venusian atmosphere, they detect and locate volcanic eruptions through infrasound remote sensing, as discussed in Section \ref{sec:approach:volcano:detection}.
Whenever a balloon locates an eruption, it adds it to the event database; when two balloons, or a balloon and the orbiter, are in contact with each other, they also compare and update their event databases.
Each time a balloon's event database is updated,  the balloon recomputes the optimal buoyancy-control policy by solving Equations \eqref{eq:optimal-value}-\eqref{eq:optimal-value-approx-and-coeffs}. 
Finally, the balloon executes the optimal policy by changing its buoyancy to attain the optimal altitude for its (estimated) latitude and longitude.

\subsection{Remarks \revI{and Limitations}}
\label{sec:approach:remarks}

A few comments are in order.

First, the approach strongly relies on knowledge of a stochastic wind model. The two key assumptions underpinning the approach are that (i) a numerical model of the Venusian atmosphere is available, and (ii) the numerical model is \emph{statistically representative} of the actual wind field on Venus, and can therefore be used to formulate a stochastic wind model. From a theoretical standpoint, this assumption results in no loss of generality; if the wind model is poorly known, a stochastic model with large uncertainty can be employed. However, the performance of the guidance approach strongly depends on the quality of the model; intuitively, if the winds are highly uncertain, the optimal policy will be unable to effectively steer balloons towards events of interest. From a practical standpoint, high-quality mesoscale models of the Venusian atmosphere are available \cite{lebonnois2010superrotation,scarica2019validation} - such models are discussed in more detail in Section \ref{sec:experiments:setup:wind}.

Second, in order to execute the optimal policy, the location of the balloons should be known with lower uncertainty compared to the stride of the lattice used to solve the MDP in Section \ref{sec:approach:mdp:adp}. This is a reasonable assumption, as high levels of positioning accuracy can be attained through RF ranging \cite{ellis2020use,CheungLeeEtAl2019} from the orbiter and from Earth. Future work will consider the use of 
 \revI{planning tools for partially observable Markov decision processes (POMDPs) such as QMDP \cite{littman1995learning}} that can capture uncertainty in the balloons' knowledge of their location.

Finally, each balloon uses \emph{all} events in the event database to compute the rewards in Equation \eqref{eq:mdp:rewards}. With this approach, multiple balloons may attempt to revisit the same eruption, rather than maximizing the number of \revII{distinct} eruptions that are visited. Given the low number of events that the balloons are likely to detect, and the strong science interest in revisiting  an eruption multiple times, we do not consider this a limitation of our approach; for scenarios where multiple revisits are not desirable, future work will consider distributed task allocation policies that assign specific eruptions to specific balloons.

\section{Numerical Experiments: Setup} %
\label{sec:experiments:setup}

We assess the performance of the proposed approach through numerical experiments. 
In our experiments, we evaluate the number of visits to volcanic events of interest for three different scenarios corresponding to different balloon guidance laws, that we call autonomous, ground-in-the-loop and passive:

\begin{itemize}
    \item \textit{autonomous:} the balloons detect events on board, and replan their altitude profile on board to visit them, according to the concept of operations described in Section \ref{sec:problem};  balloons communicate event detections with each other and with the orbiter whenever possible;
    \item \textit{ground-in-the-loop:} the balloons detect events and report them to the orbiter; the orbiter communicates with Earth twice a day at pre-specified times; Earth computes optimal policies to visit detected events (by solving Equations  \eqref{eq:optimal-value}-\eqref{eq:optimal-value-approx-and-coeffs}); the policies are uplinked to orbiter, and orbiter distributes them to balloons at the first available communication opportunity. Compared to the autonomous policy, the  ground-in-the-loop policy effectively introduces a temporal delay in implementation of the policy on the balloons.
    \item \textit{passive:} the balloons drift in the wind field, with no active buoyancy control and no particular destination.
\end{itemize}

In the rest of this section, we discuss in detail the models used for events of interest and their detection, the wind model, the balloons' and the orbiter's motion model, and the communication model.

\subsection{Volcanic Event Model} %
\label{sec:setup:volcanoes}
\subsubsection{Eruption Locations} 
\label{sec:setup:volcanoes:location}
\begin{figure}
    \centering
    \begin{subfigure}[b]{\linewidth}
    \includegraphics[width=.75\textwidth]{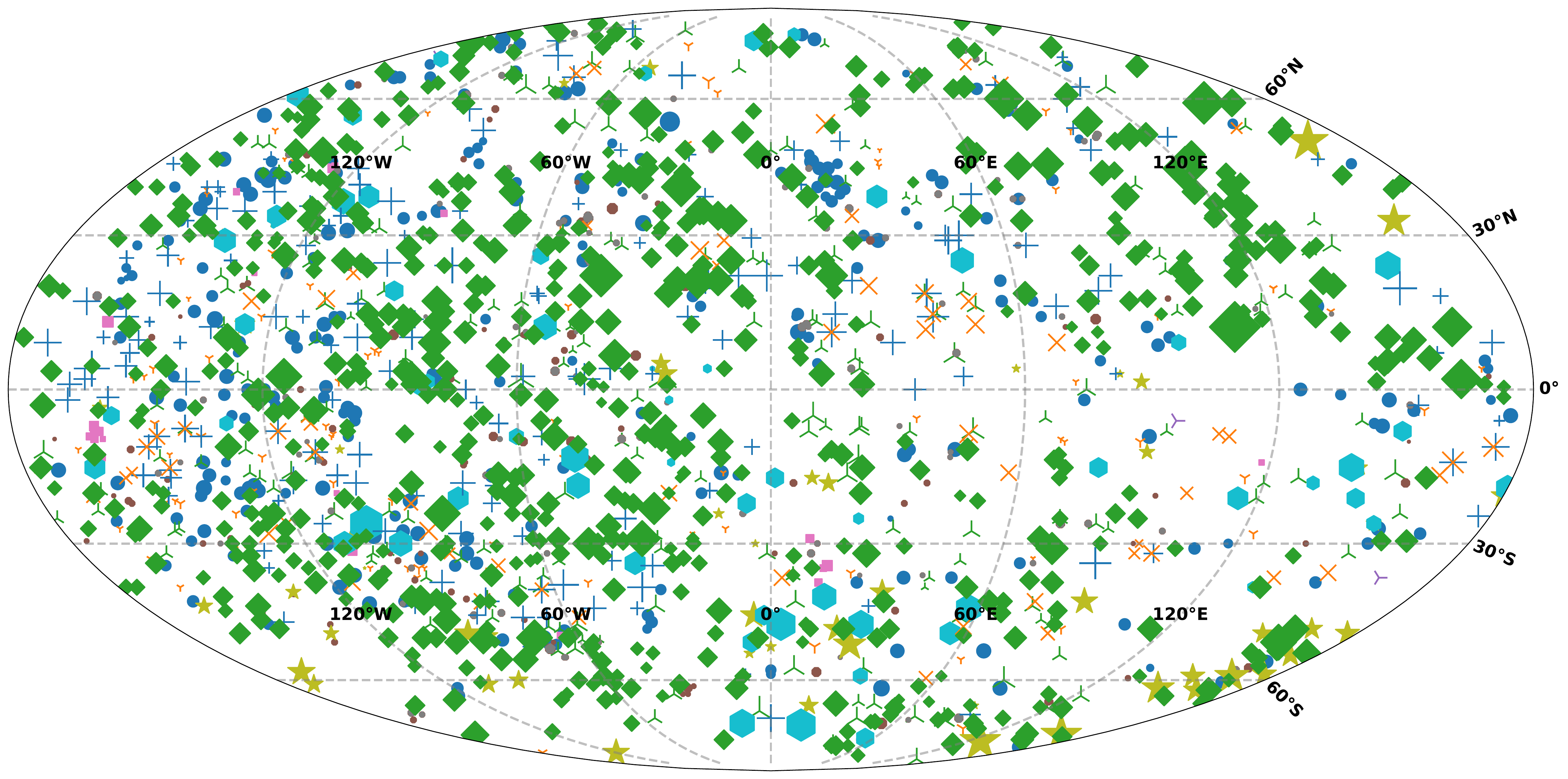} \includegraphics[width=.24\textwidth]{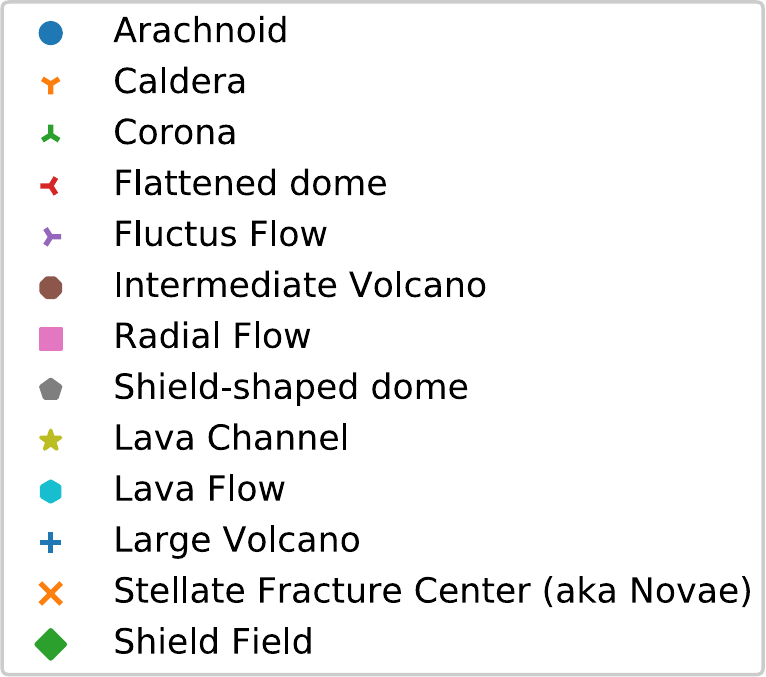}
    \caption{Location of all volcanic features identified by the Magellan spacecraft}
    \label{fig:locations:all}
    \end{subfigure}
    \begin{subfigure}[b]{\linewidth}
    \includegraphics[width=\textwidth]{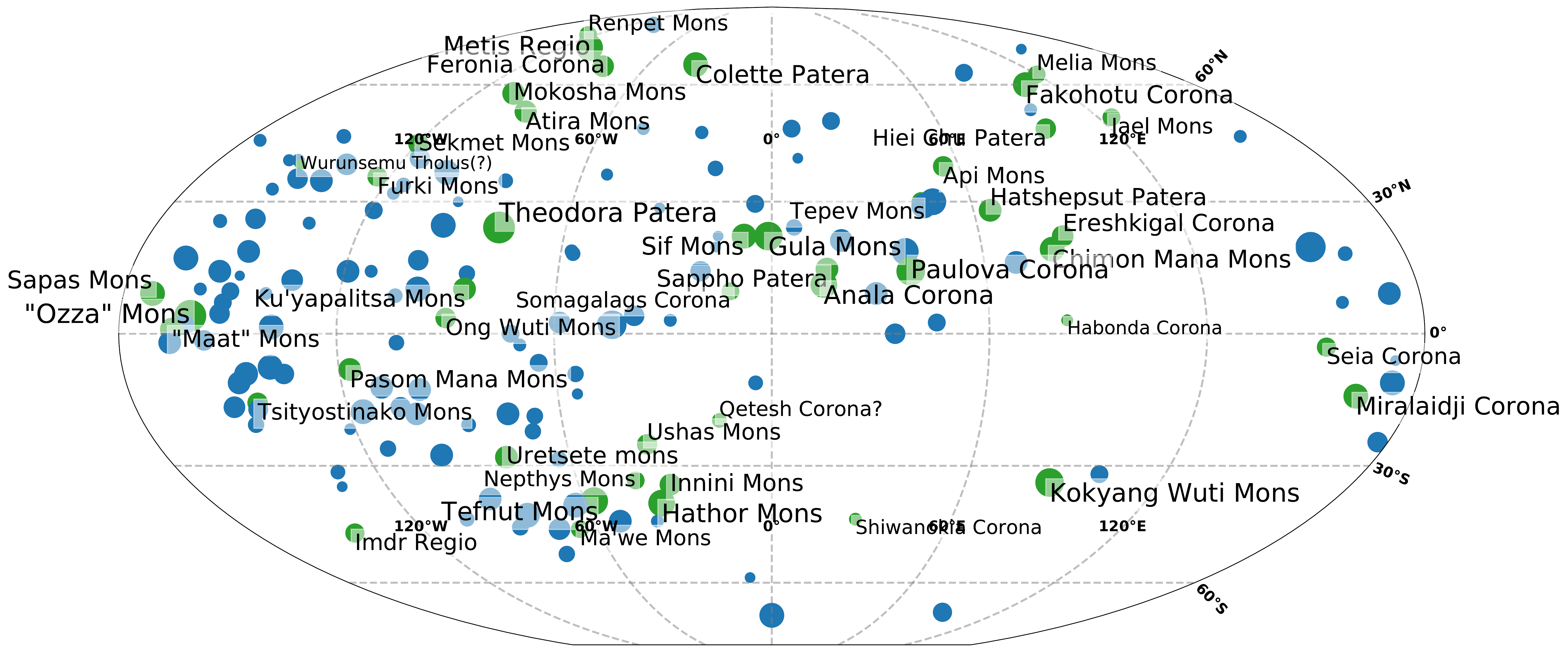}
    \caption{Location of large volcanoes identified by the Magellan spacecraft}
    \label{fig:locations:large}
    \end{subfigure}
    \caption{Volcanoes locations identified by the Magellan spacecraft \cite{CrumplerVenusCatalog}. The marker size denotes the approximate size of the volcano}
    \label{fig:locations}
\end{figure}

We used the locations and dimensions of the volcanoes identified by the Magellan spacecraft~\revI{\cite{head1992venus,CrumplerVenusCatalog}}  (Figure~\ref{fig:locations:all}), many of which are within $\pm30$ degrees latitude. %
Over a thousand volcano locations have been identified, with various sizes. Since we are interested in evaluating visits occurring after a detection, events whose detection range are very close to the visit range were not used: accordingly, we use only the locations of the 168 ``large volcanoes'' identified in the \revI{Magellan Venus volcanic feature catalog \cite{CrumplerVenusCatalog}}, shown in Figure \ref{fig:locations:large}.

\subsubsection{Eruption Type and Duration}
\label{sec:setup:volcanoes:duration}
\begin{figure}[t]
    \centering
        \begin{subfigure}[b]{.32\linewidth}
    \includegraphics[width=\textwidth]{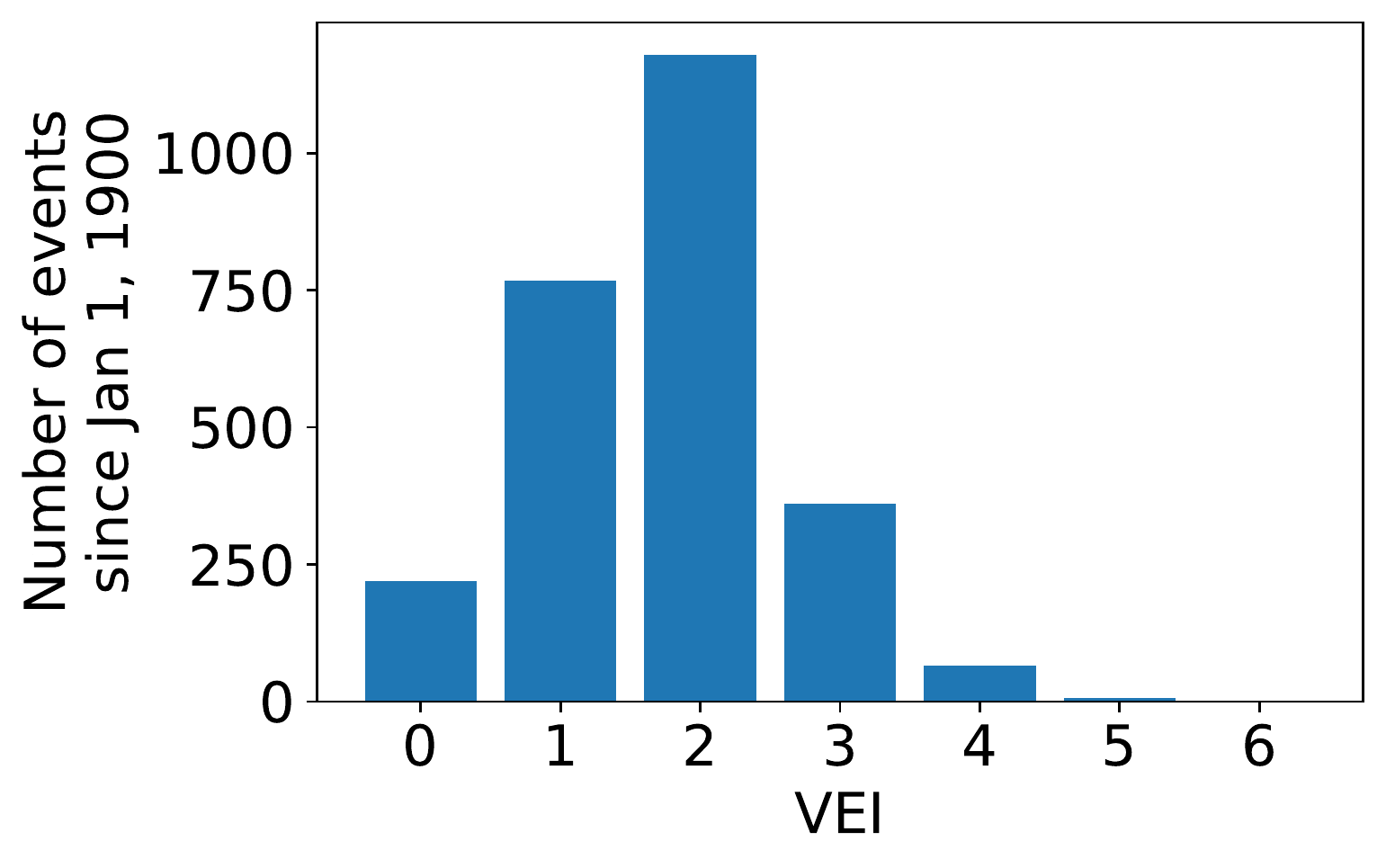}
    \caption{Eruption frequency by VEI since 1900.}
    \label{fig:earth_model:vei}
    \end{subfigure}
        \begin{subfigure}[b]{.32\linewidth}
    \includegraphics[width=\textwidth]{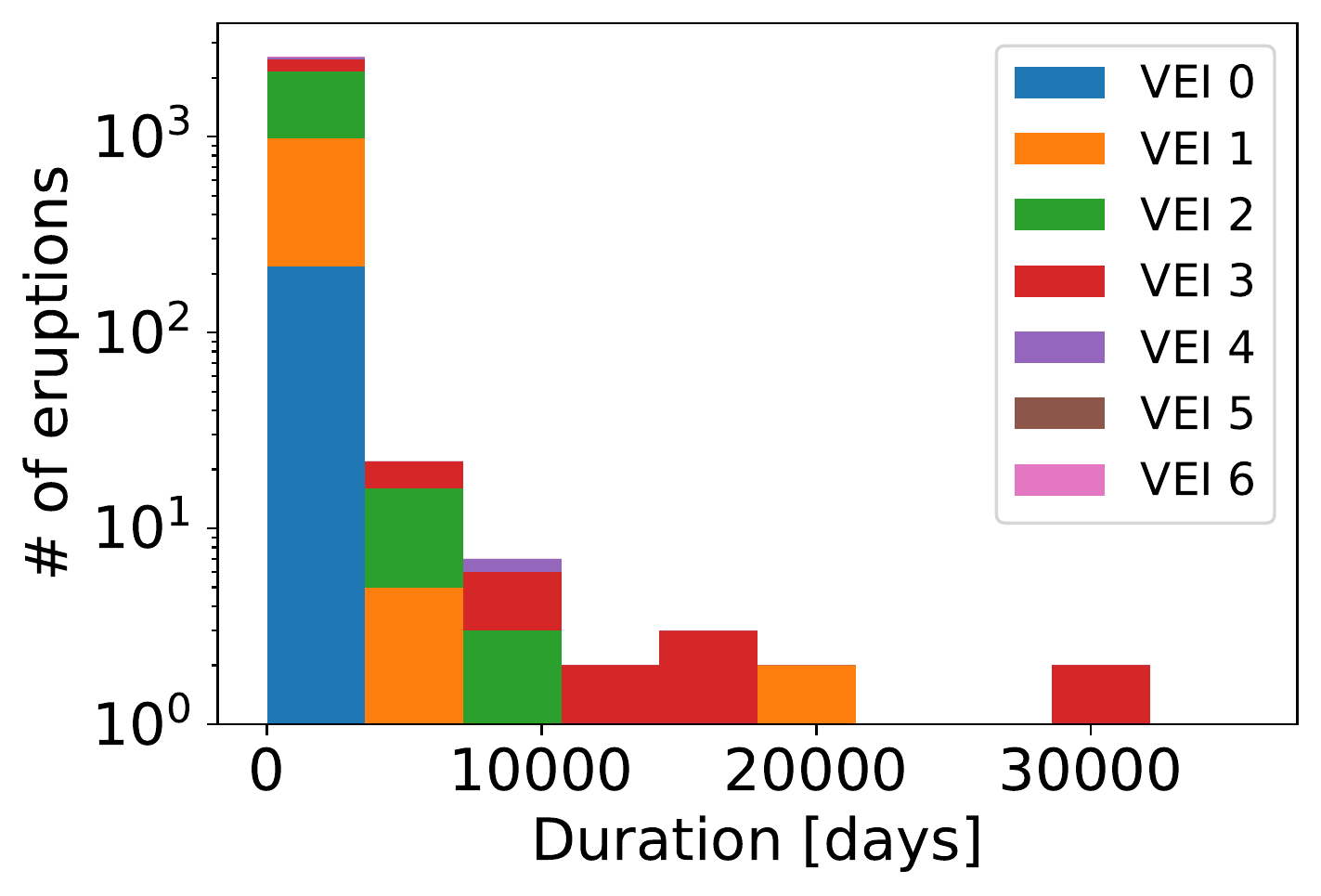}
    \caption{Duration of volcanic eruptions by VEI.}
    \label{fig:earth_model:duration}
    \end{subfigure}
        \begin{subfigure}[b]{.32\linewidth}
    \includegraphics[width=\textwidth]{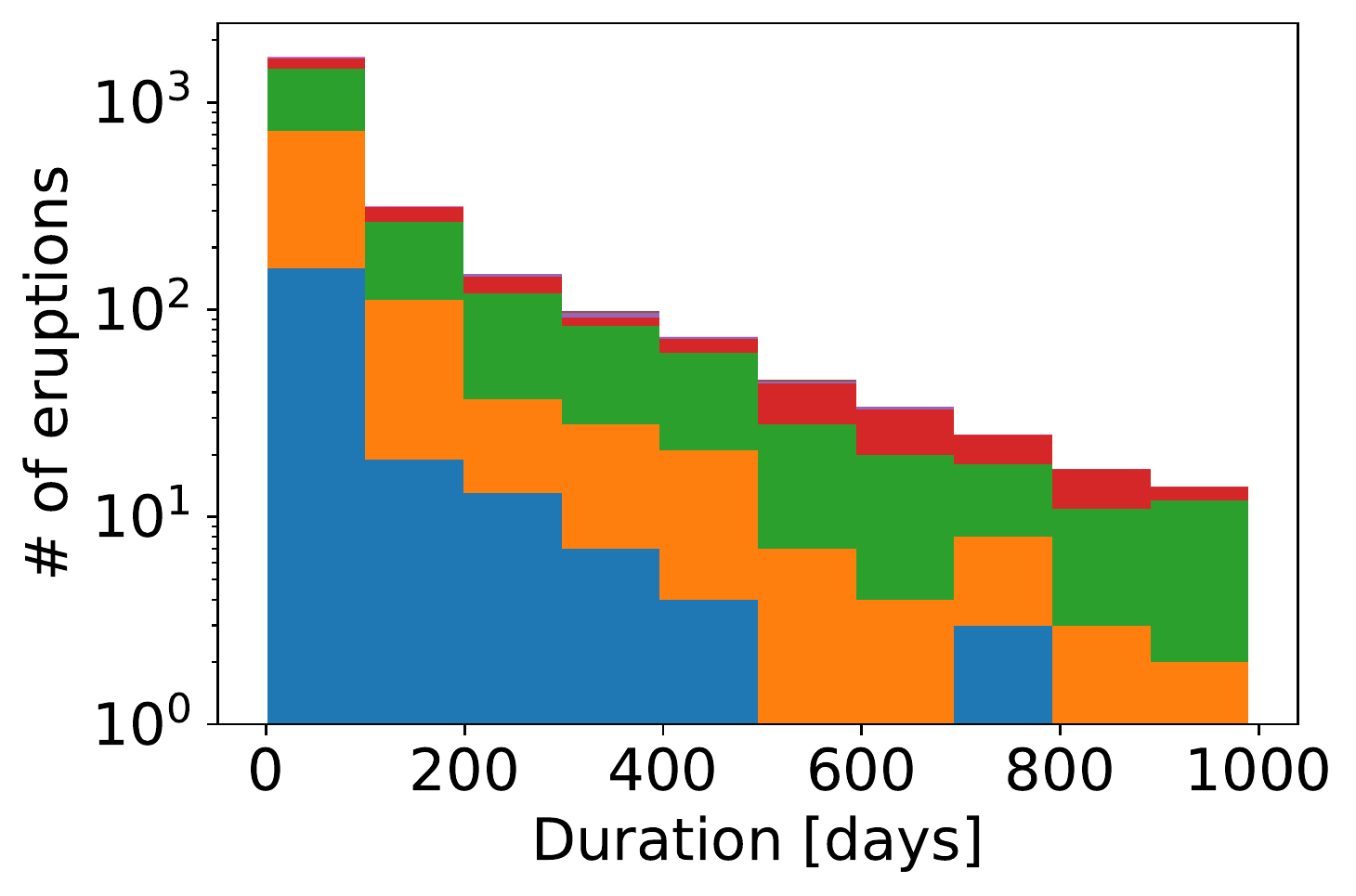}
    \caption{Duration of volcanic eruptions by VEI (detail).}
    \label{fig:earth_model:duration:detail}
    \end{subfigure}

    \caption{Eruption type and duration based on Earth-like volcanism}
    \label{fig:earth_model}
\end{figure}

Volcano explosivity is characterized by volcanic explosivity index (VEI). Because eruption type and duration in Venus are unknown, in this study we model them based on Earth-like volcanism. We chose the Smithsonian Global Volcanism program’s database \cite{VolcanoesOfTheWorld} as the repository for Earth volcanoes, and focused our analysis on eruptions recorded since Jan. 1, 1900, since older data presents a bias towards larger eruptions.
 VEI on Earth can range from 0 (Hawaii) to 8. Mt. Pinatubo (1991) had a VEI of 6. Data for the last 122 years on Earth volcanism shows that the median eruption VEI is 2 (Figure \ref{fig:earth_model:vei}). Most eruptions last less than one day, but the distribution has a long tail, with some eruptions lasting as long as 10,000 days, as shown in Figures~\ref{fig:earth_model:duration} and \ref{fig:earth_model:duration:detail}.

We chose large volcanoes on Venus as the locations of the simulated eruptions, and randomly generated eruptions at these locations with frequency and intensity drawn from the Earth volcanism dataset. 
In Section \ref{sec:experiments:results:event_frequency}, we relax this assumption by exploring the sensitivity of the approach's performance to the frequency of volcanic events through extensive numerical simulations.

\subsection{Wind and Balloon Motion Model}
\label{sec:experiments:setup:wind}

The atmospheric circulation model was generated using the state-of-the-art  Institut Pierre Simon Laplace (IPSL) Venus Global Circulation model (GCM) \cite{lebonnois2010superrotation}, recently validated in~\cite{scarica2019validation}\revI{\footnote{The Venus IPSL GCM is freely available, by request, at \url{http://www-venus.lmd.jussieu.fr/}.}}. The model is based on the \revI{Laboratoire de Météorologie Dynamique-Zoom (LMDZ)} latitude-longitude grid finite-difference dynamical core \cite{Hourdin2006}. The IPSL Venus GCM horizontal resolution is of 96 longitudes by 96 latitudes ($3.75 \times 1.875$ degrees, \revI{or $\approx 400\times 200$ km at the Equator and $\approx 200\times 200$ km at 60-degree latitude}), on 50 levels from surface to roughly 95 km altitude. The GCM produces mesoscale structures that compare well with observed circulation patterns on Venus observed by the Vega balloons and the Venus Express and Akatsuki orbiters. A reference GCM simulation spans 1 Venus day (117 Earth days); the simulation starts with the atmosphere at rest (uniform zonal wind field $u = 0$), with a vertical temperature profile very close to the \revI{Venus International Reference Atmosphere (VIRA)} model. Surface pressure (92 bar) and temperature profile are set to be the same for all surface locations, without capturing topography. \cite{garate2018latitudinal}

The output of the model includes zonal and meridional winds for every time step, latitude, longitude, and pressure level included in the model grid.

\begin{figure}[t]
    \centering
	\includegraphics[width=\textwidth]{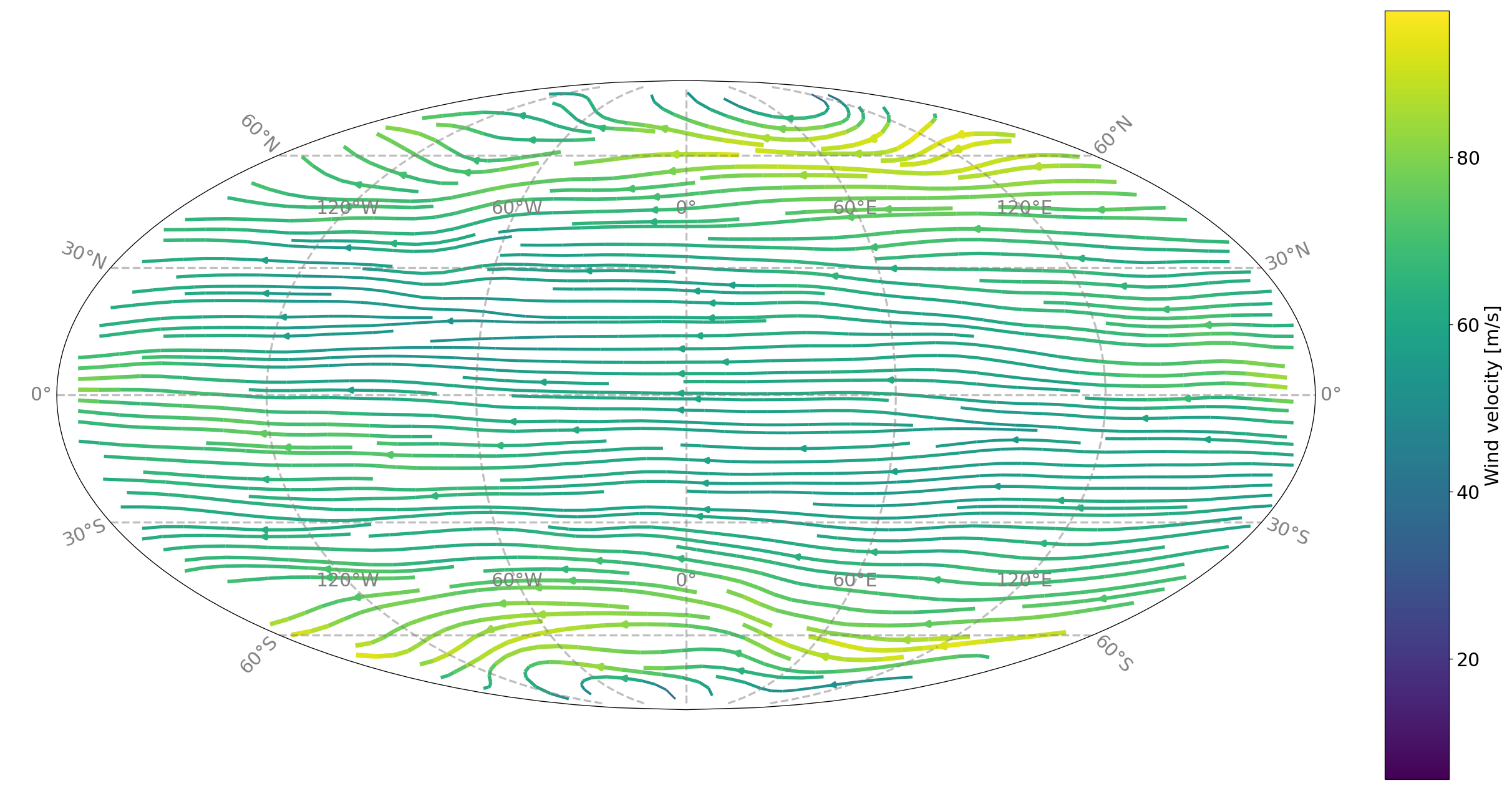}
    \caption{Wind field 58.43 days into the simulation at a pressure altitude of 27579.54 Pa (corresponding to an approximate altitude of 56 km, depending on local temperature) above the Venusian surface.}
    \label{fig:wind_field}
\end{figure}

Figure~\ref{fig:wind_field} shows the wind field for a selected time step and altitude. A movie depicting the temporal evolution of the wind field is provided in the Supplementary Material. %

\revI{Balloons were constrained to fly between minimum and maximum altitudes of $\underline h = 47 km$ and $\overline h = 63 km$.
The localization accuracy of the balloons was assumed to be better than the IPSL Venus GCM's resolution of $\approx200$ km at latitudes under $60^\circ$\revII{ --- a very loose requirement, since localization accuracy of tens of meters is achievable via ranging from an orbiter \cite{ellis2020use}}. 
\revII{
We remark that, while better localization accuracy would not immediately benefit execution of the proposed long-range guidance policy due to the low resolution of the underlying wind model, it would be highly beneficial for fine terminal guidance (e.g., \cite{bellemare2020autonomous}).
In addition, we stress that the resolution of the proposed policy is limited by the resolution of the IPSL wind model; if a higher-resolution model of atmospheric currents was available, the approach in this paper could be readily employed to derive higher-resolution guidance policies, which would in turn benefit from improved localization accuracy.
 However, our results show that, even with the comparatively low-resolution wind model considered in this work, the proposed technique is able to guide balloons to events of interests, and outperforms a passive approach.
}

In the simulations, the initial location of all three balloons was selected uniformly at random across the surface of Venus, subject to the condition that at least one balloon was within detection range of an active volcanic event at the initial time step. This condition, enforced through rejection sampling, ensured that the study would focus on the ability of the proposed guidance policy to revisit detected events (rather than detect new ones), in line with the proposed contribution of this work.
}

\revI{The MDP costs $r_{\text{altitude}}$, $r_{\text{eruption}}$, and $r_{\text{energy}}$ were set respectively  to $-10^6$, $10^3$ and $0$ (effectively assuming that the energy cost for changing altitude is dominated by the "housekeeping" energy cost to support payloads and communications). This selection of costs empirically ensured that the balloons would not exit the safe altitude ranges in the pursuit of an eruption, while also allowing them to maneuver as needed to reach an eruption.
}

\subsection{Orbiter Motion Model}

We consider two different orbits for our orbiter, and compare their relative performance. We first consider the orbit of the VAMOS orbiter concept \cite{sutin2018,didion2018}, a circular, equatorial orbit \revII{with an altitude of 30,000 km}. This orbit has a full view of half the planet at any given time, and a long orbital period of 20.91 hours. The second orbit considered is that of the planned VERITAS mission \cite{smrekar2022veritas}, a 220 km nearly polar circular orbit with a fairly restricted view of the planet, but a much faster orbital period of 1.53 hours.  

\begin{figure}[h]
    \centering
	\includegraphics[width=.8\textwidth]{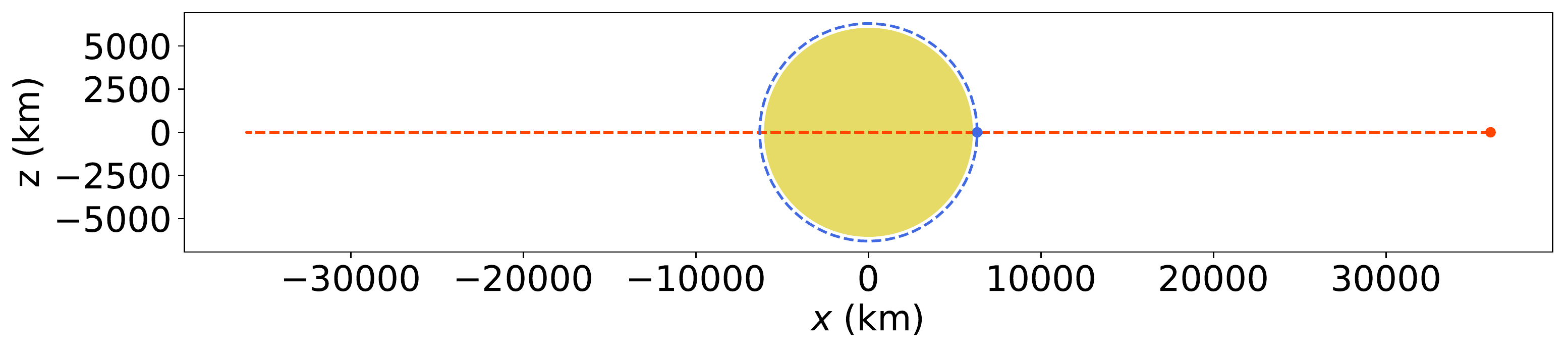}
	\includegraphics[width=.6\textwidth]{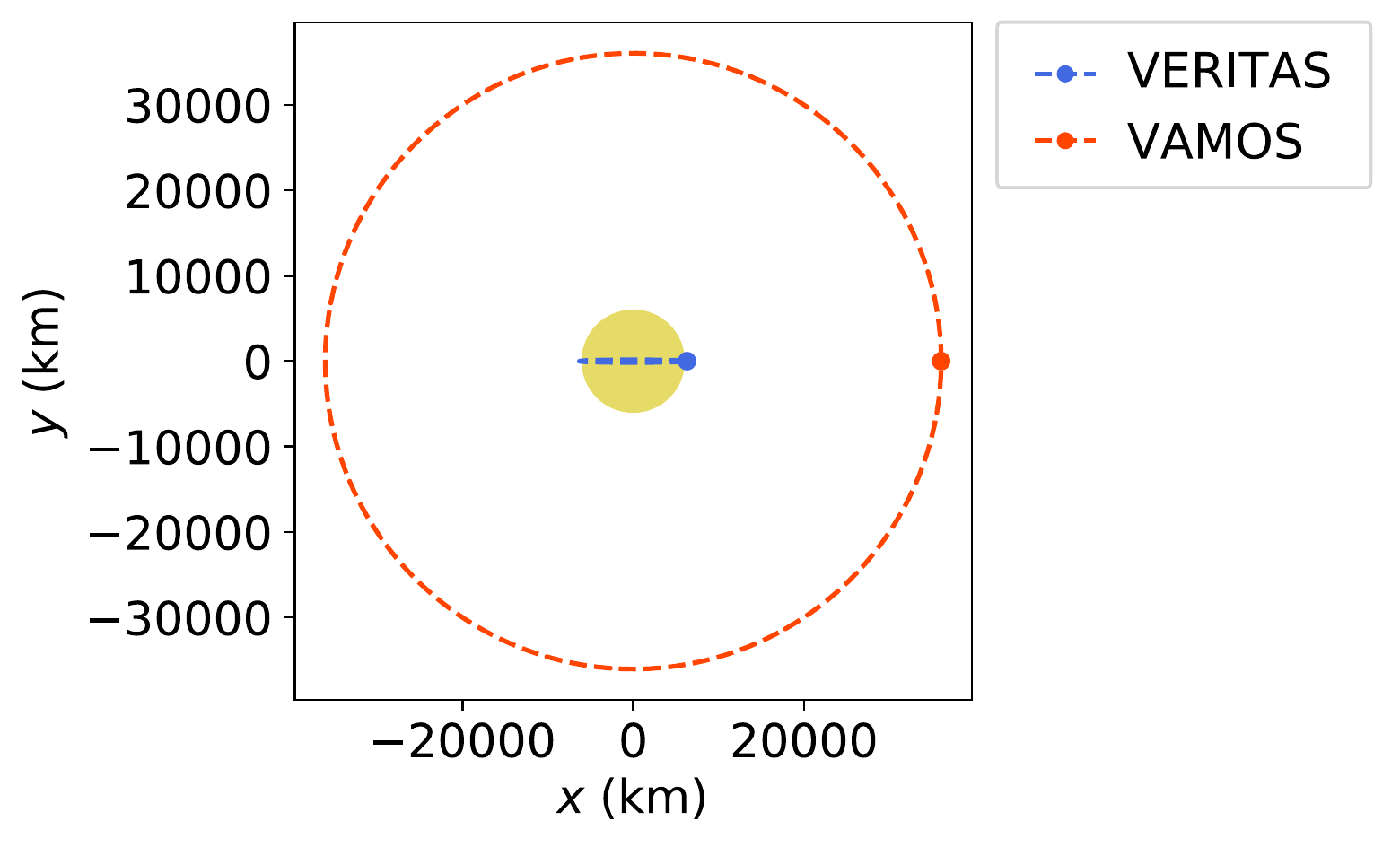}
    \caption{Orbits of the proposed VAMOS (red) and VERITAS (blue) orbiters}
    \label{fig:orbits}
\end{figure}

The orbits of the VAMOS mission concept and of the planned VERITAS mission are shown in Figure \ref{fig:orbits}.

\subsection{Communication Model}  

\begin{figure}[t]
\centering
\includegraphics[width=\textwidth]{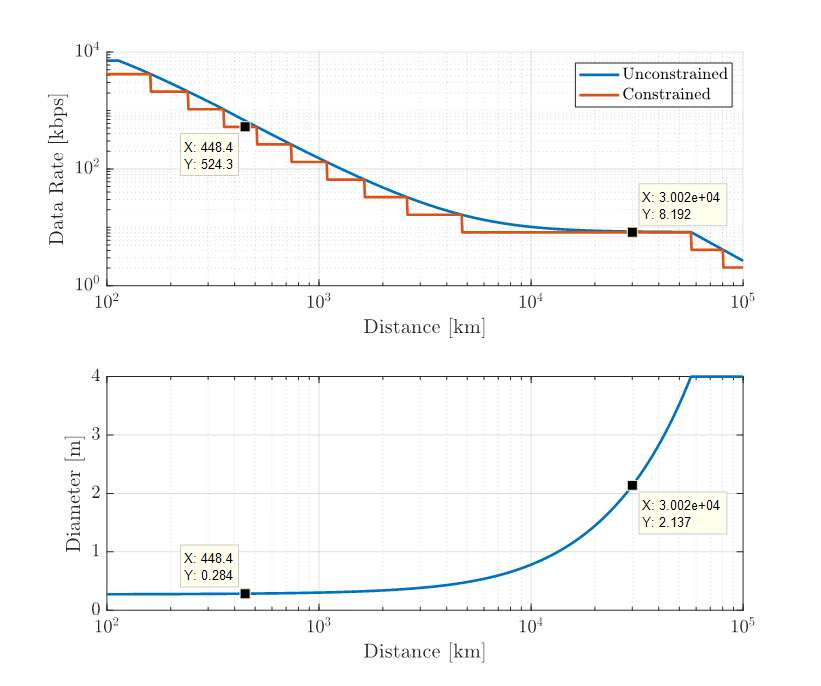}
\caption{Data rate between balloons and the orbiter as a function of range. The diameter of the orbiter's antenna is designed to cover the full surface of Venus within its -3dB beamwidth, as shown in the bottom figure. The ``constrained'' line captures the fact that, in practical applications, radios typically only operate at data rates that are power of 2.}
\label{fig:results:venus_link_budget}
\end{figure}

We consider a communication model where balloons can exchange information with each other whenever they are within 200 km range of each other; and balloons can communicate with the orbiter whenever the orbiter is at least $30^\circ$ above the horizon (irrespective of range). We assumed that  the balloons are equipped with dipole antennas with 2dB of gain and a power budget of 1W; the orbiter is equipped with a directional antenna that covers the entire disc of Venus within the antenna's -3dB beamwidth; the blackbody temperature of antennas on Venus is 730 K \cite{DSN810005}; and system losses account for 3dB of attenuation. With these parameters, the VAMOS orbiter is able to communicate with the balloons at 8 kbps, and the VERITAS orbiter is able to communicate at over 500 kbps (albeit much less frequently), as shown in Figure \ref{fig:results:venus_link_budget}; between the balloons, data rates of up to 250 kbps can be achieved.

\section{Numerical Experiments: Results}
\label{sec:experiments:results}

We are now in a position to assess the performance of the proposed guidance approach. We investigate \revI{five} key questions:
\begin{itemize}
\item \textbf{Effect of on-board autonomy}: Does on-board autonomous detection and guidance result in an increased number of events visited, and in a reduction in the time between detection and visit, compared to \revI{ground-in-the-loop guidance and to uncontrolled drifters (passive guidance)}?
\revI{
\item \textbf{Sensitivity to orbiter's orbit}: How does the performance of the approach depend on the orbiter's orbit, which affects both the availability of opportunities for inter-balloon communication, and the probability of detecting volcanic events?
}
\item \textbf{Sensitivity to detection radius}: How does the performance of the approach depend on the distance at which volcanic events can be detected?
\item \textbf{Sensitivity to event frequency}: How does the approach perform in cases where the volcanic events are less frequent, or more frequent, than expected?
\item \textbf{Sensitivity to fleet size}: How does the approach's performance scale with the number of balloons in the fleet? 
\end{itemize}

To address these questions, we performed a thousand simulations for each of the scenarios of interest described below. Each simulation spans 60 Earth days. For each simulation, the locations, VEIs, and durations of volcanic events are randomly sampled as described in Section \ref{sec:setup:volcanoes}. The initial locations of the balloons are randomly selected (in order to focus our investigation on the ability of the approach to revisit detected events, the sampling process ensures that at least one eruption is within detection range of one balloon at the initial time step). The balloons' locations are propagated as described in Section \ref{sec:experiments:setup:wind}. The starting time for the wind field is randomly selected among all time steps captured by the numerical atmospheric circulation model, and the wind field is then propagated for 60 Earth days. If the simulation time exceeds the temporal boundaries of the numerical circulation model, the model rolls over to the first time step.

Figure \ref{fig:experiments:results:sim} shows the output of a representative simulation. Videos showing representative simulations for the autonomous, ground-in-the-loop, and passive cases are available in the Supplementary Material.

\revI{
\begin{figure}[h]
\centering
\includegraphics[width=\textwidth]{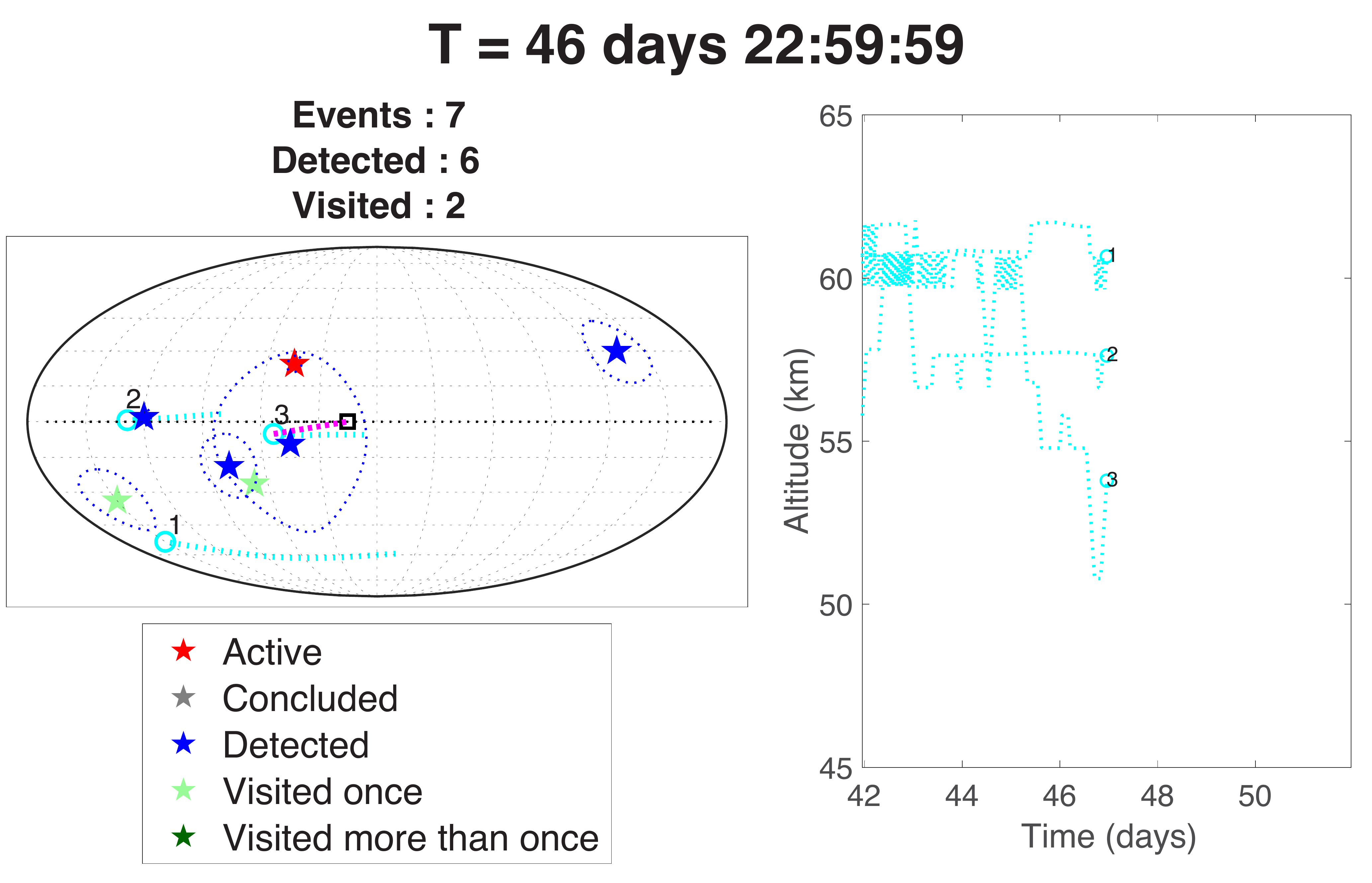}
\caption{Output of a representative simulation with on-board autonomy. Balloons are shown as cyan circles. Cyan dotted lines show the balloons' ground tracks in the previous 24 hours. The orbiter's sub-spacecraft point is shown as a black square, and the orbiter's equatorial ground track is shown as a dotted black line. Magenta lines denote active communication links between agents. Volcanic events are shown as stars; the color of the star denotes whether an event is active and whether it has been detected or visited. Blue dotted circles around the events denote the detection radius.}
\label{fig:experiments:results:sim}
\end{figure}
}

\subsection{Effect of on-board autonomy}
\label{sec:experiment:results:autonomy}

\begin{figure}[h]
\centering
\includegraphics[width=\textwidth]{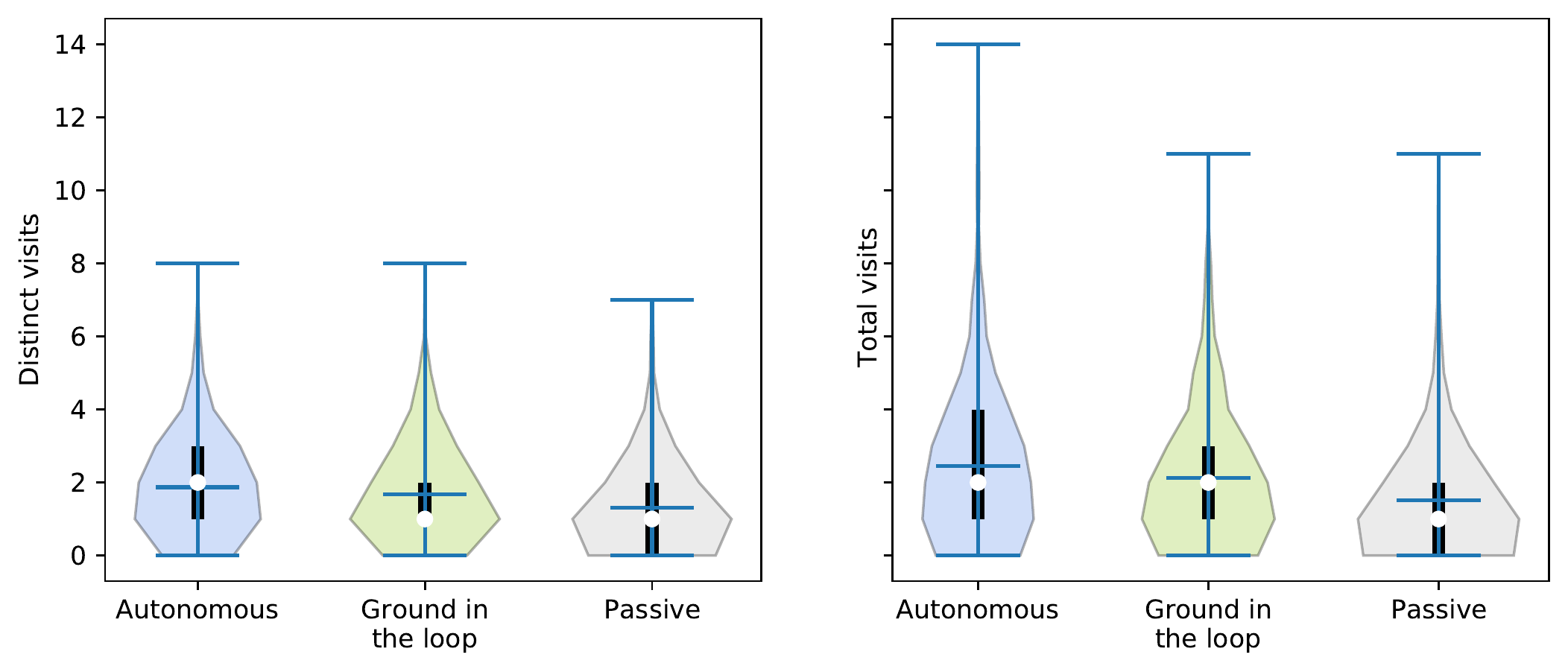}
\caption{Effect of on-board autonomy with a VAMOS orbit. The violin plot \revII{\cite{hintze1998violin}} shows the \revII{empirical} distribution of the number of distinct and total visits over 1000 trials. \revII{A vertically-aligned histogram shows the distribution of visits; the histogram is shaded in blue for the autonomous guidance controller, in green for the ground-in-the-loop controller, and in gray for the passive approach}. The white dot denotes the median; black bars denote the 25\% and 75\% quantiles; and the blue dashes denote the mean, maximum and minimum values of the distribution. \revI{The same vertical scale is used for both plots}. For each simulation, the locations, types, and durations of volcanic events are randomly selected as described in Section \ref{sec:setup:volcanoes}. }
\label{fig:results:detections:VAMOS}
\end{figure}

Table \ref{tab:results:detections:VAMOS} and Figures \ref{fig:results:detections:VAMOS} \revII{and \ref{fig:results:distances}a} report the performance of the proposed approach, compared to ground-in-the-loop autonomy and to a passive approach, for an orbit corresponding to the VAMOS orbiter concept.

\begin{table}[h]
\caption{Effect of on-board autonomy, VAMOS orbit}
\label{tab:results:detections:VAMOS}
\begin{adjustbox}{center}
\begin{tabular}{lcccccc}
\toprule
                    &  \multicolumn{2}{c}{Autonomous}  &  \multicolumn{2}{c}{Ground in the loop} &  \multicolumn{2}{c}{Passive} \\
                   &        mean &   std. dev. &                mean &           std. dev. &     mean &  std. dev. \\
\midrule
Distinct detections                &        8.35 &        3.39 &                8.33 &                3.38 &     \textbf{8.45} &       3.44 \\
Distinct visits                    &        \textbf{1.87} &        1.37 &                1.68 &                1.35 &     1.31 &       1.21 \\
Total visits                       &        \textbf{2.46} &        1.98 &                2.12 &                1.82 &     1.51 &       1.51 \\
Detected events visited [\%]        &       \textbf{23.54} &       17.91 &               21.42 &               17.70 &    15.91 &      15.02 \\
Events visited [\%]                 &       \textbf{14.24} &       11.18 &               12.85 &               10.78 &     9.87 &       9.94 \\
\bottomrule
\end{tabular}

\end{adjustbox}
\end{table}

The use of \revI{autonomous} buoyancy control has a \revI{significant} impact on the number of distinct and total events visited. Compared to the passive case, on-board autonomy results in a 42.7\% mean increase in the number of \revII{distinct} events visited \revII{(corresponding to 0.56 additional expected distinct visits)} and a 62.9\% increase in the number of total events visited \revII{(or 0.95 additional total expected visits)}. Ground-in-the-loop guidance results in a smaller, but still significant, 28.2\% increase in the number of \revII{distinct} events visited \revII{(corresponding to an extra 0.37 expected visits)}, and 40.4\% increase in the total number of visits \revII{(0.61 additional expected total visits)}, compared to the passive case.

\revII{
Figure \ref{fig:results:distances}a shows the distribution of distances between the visiting balloon and the event at the time of closest approach. The mean distance at the time of a visit is 29 km; 1.16\% of all visits overfly the event at sub-km range, and 13.1\% of visits come within 10 km of the event. While the distribution of distances is qualitatively similar for all three modes, on-board autonomy results in a larger number of visits at any given distance.

\begin{figure}[h]
    \centering
	\includegraphics[width=.8\textwidth]{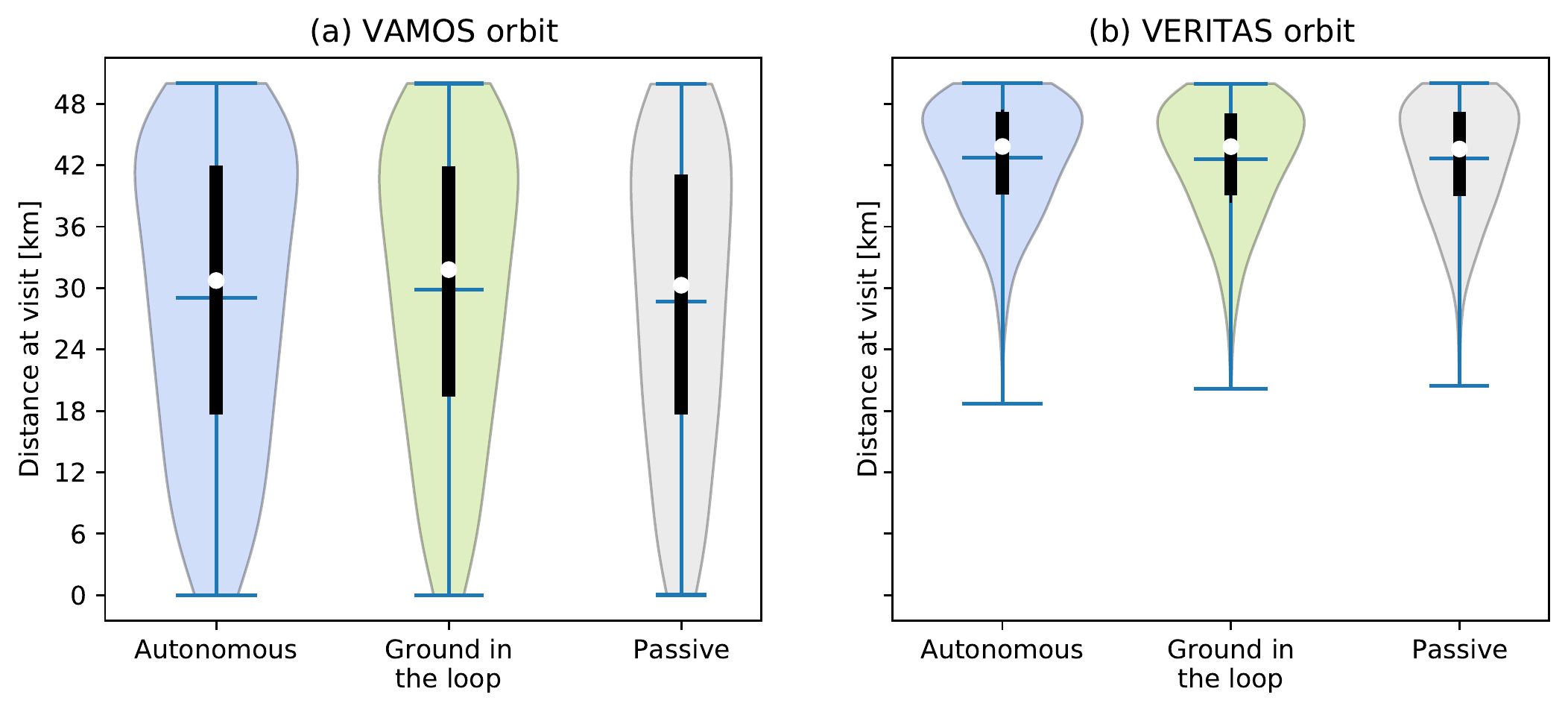}
    \caption{Horizontal distance between the balloon and the visited event at the time of closest approach.}
    \label{fig:results:distances}
\end{figure}
}
These results show that autonomous guidance (whether on-board or from the ground) holds promise to \revII{significantly} increase science returns for networked balloons investigating Venus's geology, increasing the expected number of volcanic events visited by 40\% to 60\%; and that, while on-board autonomy is required to fully realize the benefits of the proposed approach, ground-based guidance can nevertheless yield significant benefits compared to a passive approach.

\revII{While the absolute increase in the number of visited events is comparatively small, the impact on our understanding of Venus's geological phenomena is likely to be significant. On-board guidance holds promise to reduce the likelihood of observing no eruption at all from 26.7\% to 13.8\% compared to the passive case, almost halving the probability of failing to collect close-up measurements on Venus's volcanic activity.
In addition,  on-board guidance increases the likelihood of observing two or more distinct eruptions from 36.7\% to 58.2\%.
As discussed above, given the likely diversity of volcanoes and volcanic eruptions in terms of, e.g., eruptive style and material production, the ability to observe distinct volcanoes is of immense value: observing even just two distinct eruptions will provide critical information about the distribution of volcanic explosivity, the volume of ejecta, and regional variations in volcanic activity, key information to improve our understanding of Venus's geological setting.
}

\begin{figure}[h]
\centering
\includegraphics[width=\textwidth]{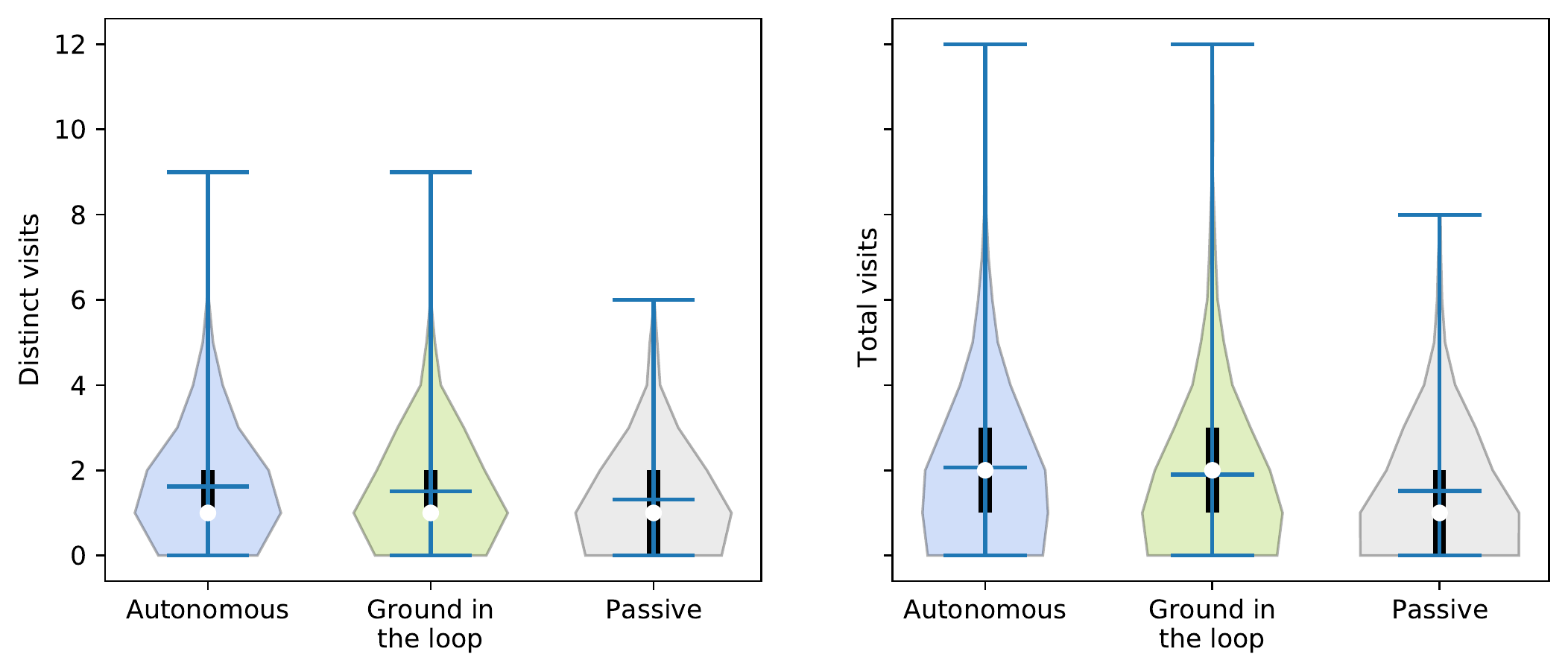}
\caption{Effect of orbiter's orbit: performance of the proposed approach with an orbit similar to the proposed VERITAS mission's.}
\label{fig:results:detections:VERITAS}
\end{figure}

\revI{
\subsection{Effect of orbiter's orbit}
\label{sec:experiment:results:orbiter}

Table \ref{tab:results:detections:VERITAS} and \revII{Figures \ref{fig:results:distances}b and} \ref{fig:results:detections:VERITAS} report the performance of the proposed approach for an orbit corresponding to the VERITAS mission concept, with significantly lower altitude and shorter orbital period compared to VAMOS's orbit.}

\begin{table}
\caption{Effect of on-board autonomy, VERITAS orbit}
\label{tab:results:detections:VERITAS}
\begin{adjustbox}{center}
\begin{tabular}{lcccccc}
\toprule
                    &  \multicolumn{2}{c}{Autonomous}  &  \multicolumn{2}{c}{Ground in the loop} &  \multicolumn{2}{c}{Passive} \\
                   &        mean &   std. dev. &                mean &           std. dev. &     mean &  std. dev. \\
\midrule
Distinct detections                &        6.78 &        3.06 &                6.80 &                2.99 &     \textbf{7.00} &       3.05 \\
Distinct visits                    &        \textbf{1.62} &        1.32 &                1.51 &                1.28 &     1.31 &       1.19 \\
Total visits                       &        \textbf{2.06} &        1.80 &                1.90 &                1.79 &     1.52 &       1.46 \\
Detected events visited [\%]        &       \textbf{24.15} &       18.85 &               23.04 &               19.60 &    18.86 &      16.79 \\
Events visited [\%]                 &       \textbf{12.08} &        9.74 &               11.51 &               10.32 &     9.57 &       8.42 \\
\bottomrule
\end{tabular}
\end{adjustbox}
\end{table}

The VERITAS orbit results in slightly worse performance compared to the VAMOS orbit: the lower orbit provides both fewer opportunities to detect events from orbit and, critically, fewer opportunities for the balloons to communicate with the orbiter. \revII{
The distance to the event at the time of closest approach is also significantly increased:  the mean distance of a visit increases to 42.8 km, and no balloon approaches the event closer than 18 km.
}
Nevertheless, on-board autonomy results in a 23.6\% increase in the number of \revII{distinct} events visited, and ground-in-the-loop guidance provides a 15.3\% increase, compared to the passive case.

\revI{Thus, while a high-altitude orbit offers better performance in terms of event detection, the proposed concept can also accommodate lower-altitude orbiters, which may be better suited for complementary scientific investigations.
}

\FloatBarrier
\subsection{Sensitivity to detection radius}
\label{sec:results:detection_radius}

\begin{figure}[h]
\centering
\includegraphics[width=\textwidth]{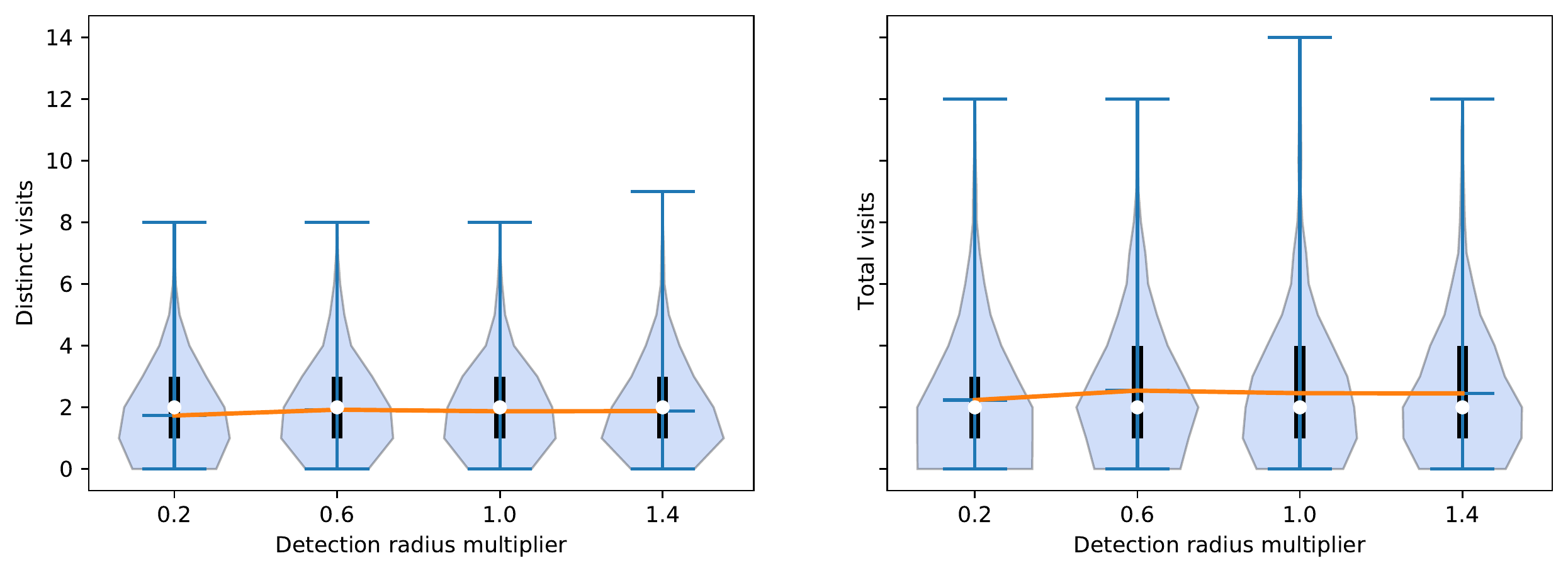}
\caption{Sensitivity of the approach's performance to the eruption detection radius}
\label{fig:results:detection_radius}
\end{figure}

\begin{figure}[h]
\centering
\includegraphics[width=\textwidth]{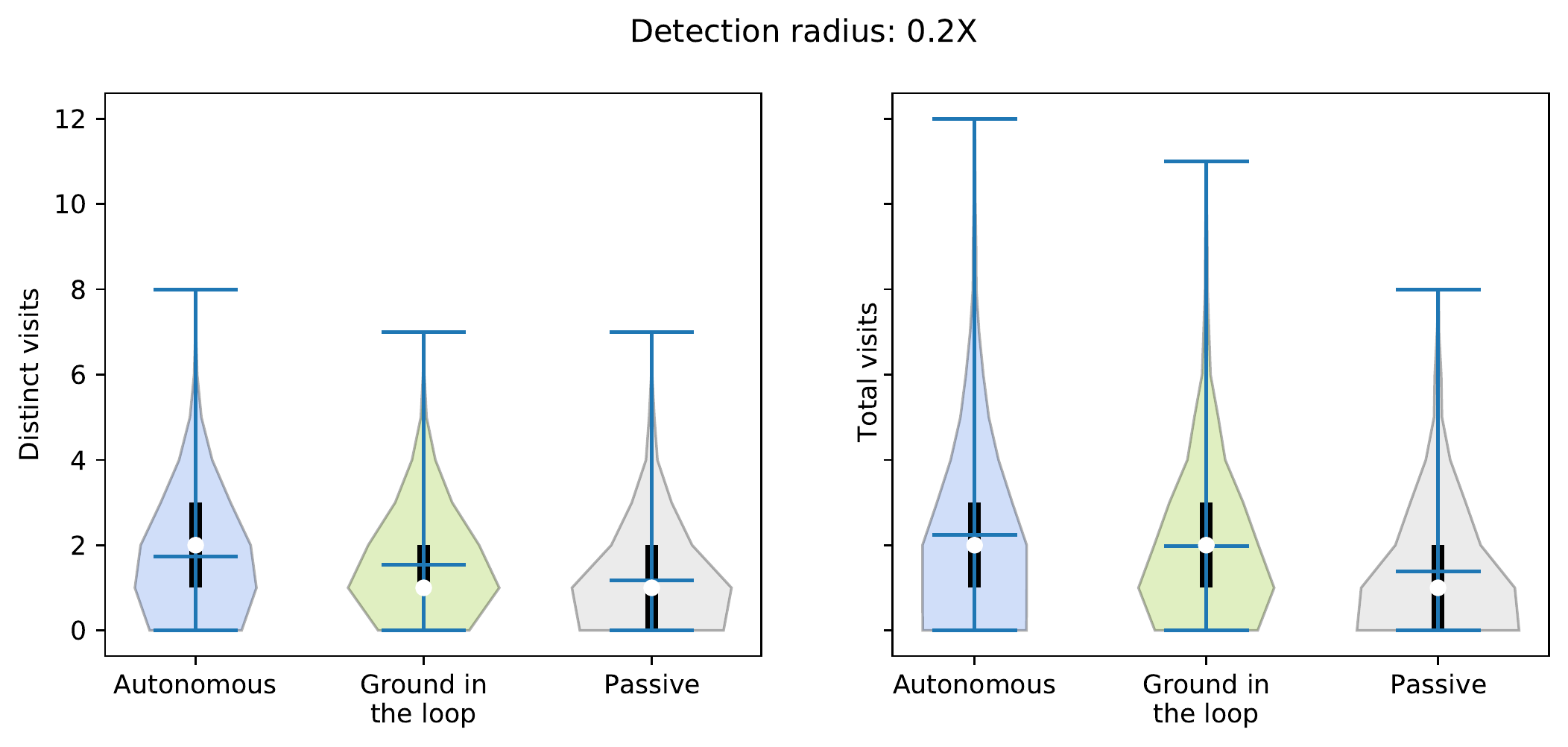}
\caption{Performance of the proposed approach with an 80\% reduction in the eruption detection radius}
\label{fig:results:detection_radius:compare}
\end{figure}

 To account for this uncertainty, we assessed the sensitivity of the proposed approach to the detection radius of the balloon's detectors. To this end, we considered detection radii of 20\%, 60\%, and 140\% of the nominal detection radius reported in Table \ref{tab:model:detection-radius}, and a VAMOS-type orbiter. \revI{Results are shown in Figures \ref{fig:results:detection_radius} and \ref{fig:results:detection_radius:compare} and in Table \ref{tab:results:detection_radius:comparison}.} %

Remarkably, the number of distinct and \revII{total} visits appears to be relatively insensitive to the detection radius, with a comparatively small 8.6\% decrease in the number of of distinct visits, and a 9.14\% decrease in the number of total visits, when the detection radius is lowered from 140\% to just 20\% of the nominal one. These results suggest that the proposed approach is highly robust to uncertainty in the detection model, and that on-board autonomy provides a significant competitive advantage across a range of possible detection sensitivities.

\revI{Table \ref{tab:results:detection_radius:comparison} shows that the advantage of on-board autonomy holds across different detection radii: on-board autonomy consistently achieves \revII{56\% to 63\% more total visits}, and between 38\% and 46\% more \revII{distinct} visits, compared to passive drifters. Compared to ground-in-the-loop guidance, on-board autonomy retains a smaller, but still significant, advantage, with 16\% to 19\% more distinct visits and 12\% to 17\% more total visits.}

\begin{figure}[h]
\centering
\includegraphics[width=\textwidth]{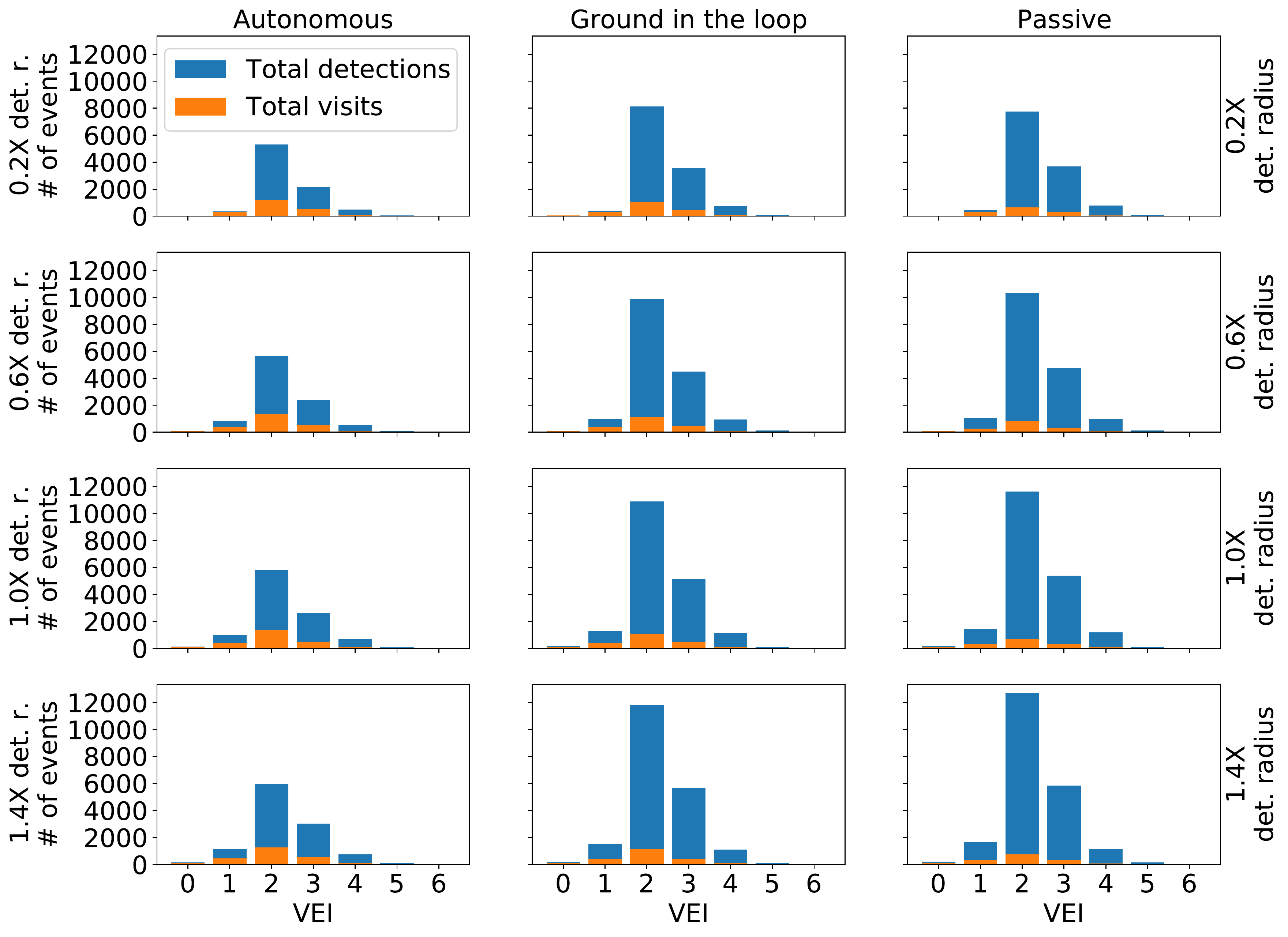}
\caption{Total number of events detected and visited as a function of \revI{detection radius and} VEI for autonomous, ground in the loop, and passive guidance modes. }
\label{fig:results:detection_radius:by_vei}
\end{figure}

 Figure \ref{fig:results:detection_radius:by_vei} explores this phenomenon in further detail by showing the number of events visited as a function of VEI for all considered detection radii and guidance laws.
 Surprisingly, as the detection radius increases, autonomy only sees a modest increase in the number of events detected, and a minimal increase in the number of events visited.
 In contrast, ground-in-the-loop and passive modes see large increases in the number of detections - since these modes are less effective at visiting events, they spend more time roaming the atmosphere with no specific target, resulting in additional serendipitous detections.
 However, the increased number of detection only translates in a minimal increase in the number of visits.
 This suggests that, as long as some events are detected, the bottleneck factor limiting the number of visits is not the number of detected events, but rather the ability to steer the balloons in Venus's atmosphere - which is exactly the issue addressed by the approach in this paper.

\begin{table}[h]
\caption{Performance of the proposed approach with a detection radius of 0.2X, 0.6X, and 1.4X of the nominal one. The nominal case is shown in Table \ref{tab:results:detections:VAMOS}.}
\label{tab:results:detection_radius:comparison}
\begin{adjustbox}{center}
\begin{tabular}{rlllllll}
\toprule
\midrule
Detection &                               &  \multicolumn{2}{c}{Autonomous} &  \multicolumn{2}{c}{Ground in the loop} &  \multicolumn{2}{c}{Passive} \\
Radius&                          &        mean &   std. dev. &                mean &           std. dev. &     mean &  std. dev. \\
\multirow{5}{*}{\rotatebox[origin=c]{90}{0.2X}} &Distinct detections                &        7.50 &        3.05 &                \textbf{7.52} &                3.02 &     \textbf{7.52} &       3.04 \\
&Distinct visits                    &        \textbf{1.73} &        1.38 &                1.55 &                1.24 &     1.18 &       1.16 \\  %
&Total visits                       &        \textbf{2.24} &        1.97 &                1.98 &                1.79 &     1.39 &       1.45 \\%13,61
&Detected events visited [\%]        &       \textbf{23.38} &       18.13 &               21.02 &               16.74 &    15.62 &      15.12 \\%
&Events visited [\%]                 &       \textbf{12.77} &       10.21 &               11.47 &                9.21 &     8.61 &       8.46 \\%
\midrule
\multirow{5}{*}{\rotatebox[origin=c]{90}{0.6X}} &Distinct detections                &        8.08 &        3.14 &                8.07 &                3.14 &     \textbf{8.13} &       3.14 \\%
&Distinct visits                    &        \textbf{1.92} &        1.44 &                1.65 &                1.32 &     1.32 &       1.20 \\%16,45
&Total visits                       &        \textbf{2.54} &        2.02 &                2.17 &                1.90 &     1.56 &       1.54 \\%17,63
&Detected events visited [\%]        &       \textbf{23.90} &       16.95 &               20.92 &               16.85 &    16.39 &      14.76 \\%
&Events visited [\%]                 &       \textbf{14.19} &       10.55 &               12.38 &               10.11 &     9.83 &       9.08 \\%
\midrule
\multirow{5}{*}{\rotatebox[origin=c]{90}{1.4X}}&Distinct detections                &        8.46 &        3.44 &                8.41 &                3.42 &     \textbf{8.52} &       3.45 \\%
&Distinct visits                    &        \textbf{1.88} &        1.44 &                1.68 &                1.34 &     1.36 &       1.26 \\%19,38
&Total visits                       &        \textbf{2.45} &        2.02 &                2.18 &                1.94 &     1.57 &       1.53 \\%12,80
&Detected events visited [\%]        &       \textbf{22.68} &       16.53 &               20.28 &               15.83 &    16.15 &      14.23 \\%
&Events visited [\%]                 &       \textbf{14.11} &       10.74 &               12.47 &                9.93 &    10.01 &       9.07 \\%
\bottomrule

\end{tabular}
\end{adjustbox}
\end{table}

\FloatBarrier
\subsection{Sensitivity to event frequency}
\label{sec:experiments:results:event_frequency}
\begin{figure}[h]
\centering
\includegraphics[width=.7\textwidth]{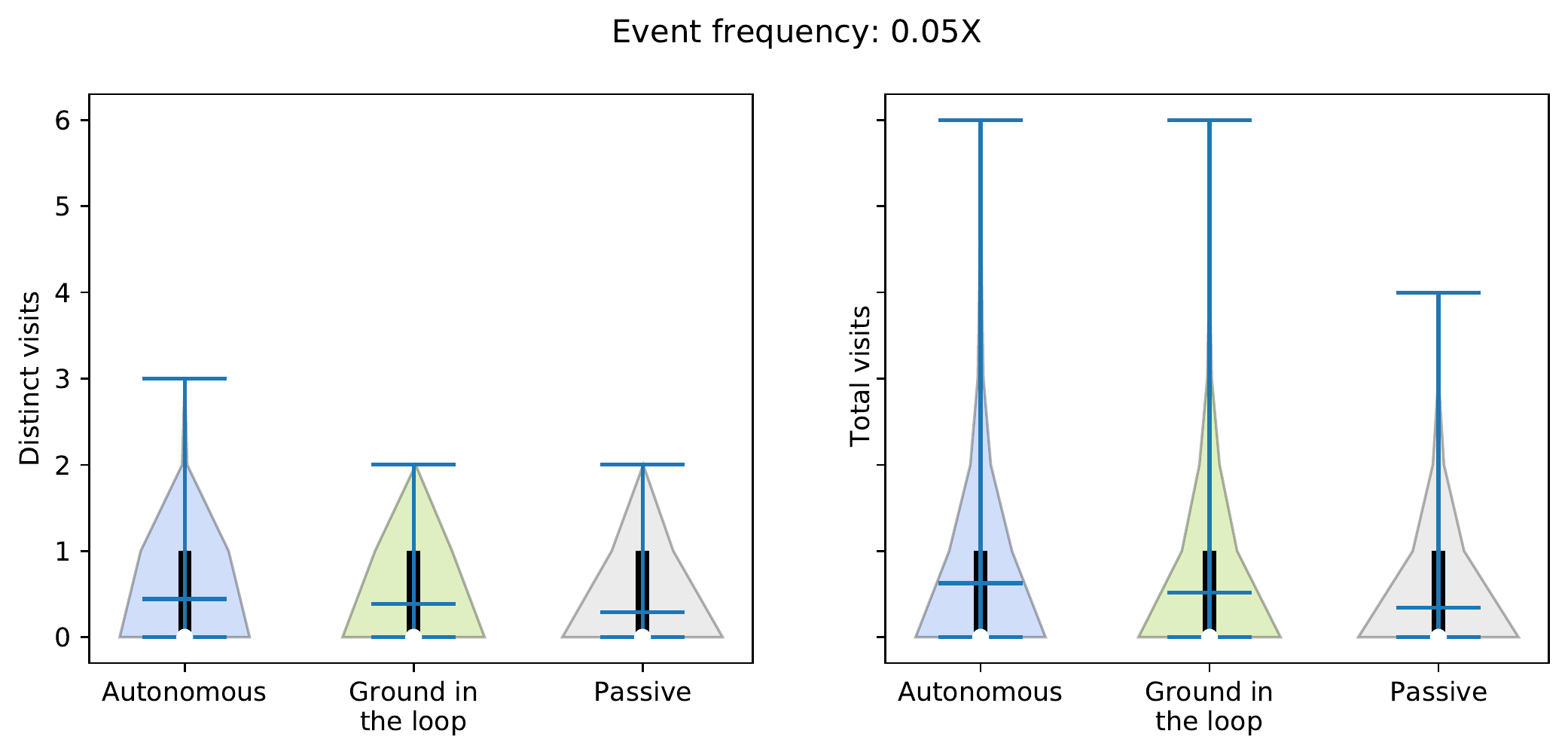}
\includegraphics[width=.7\textwidth]{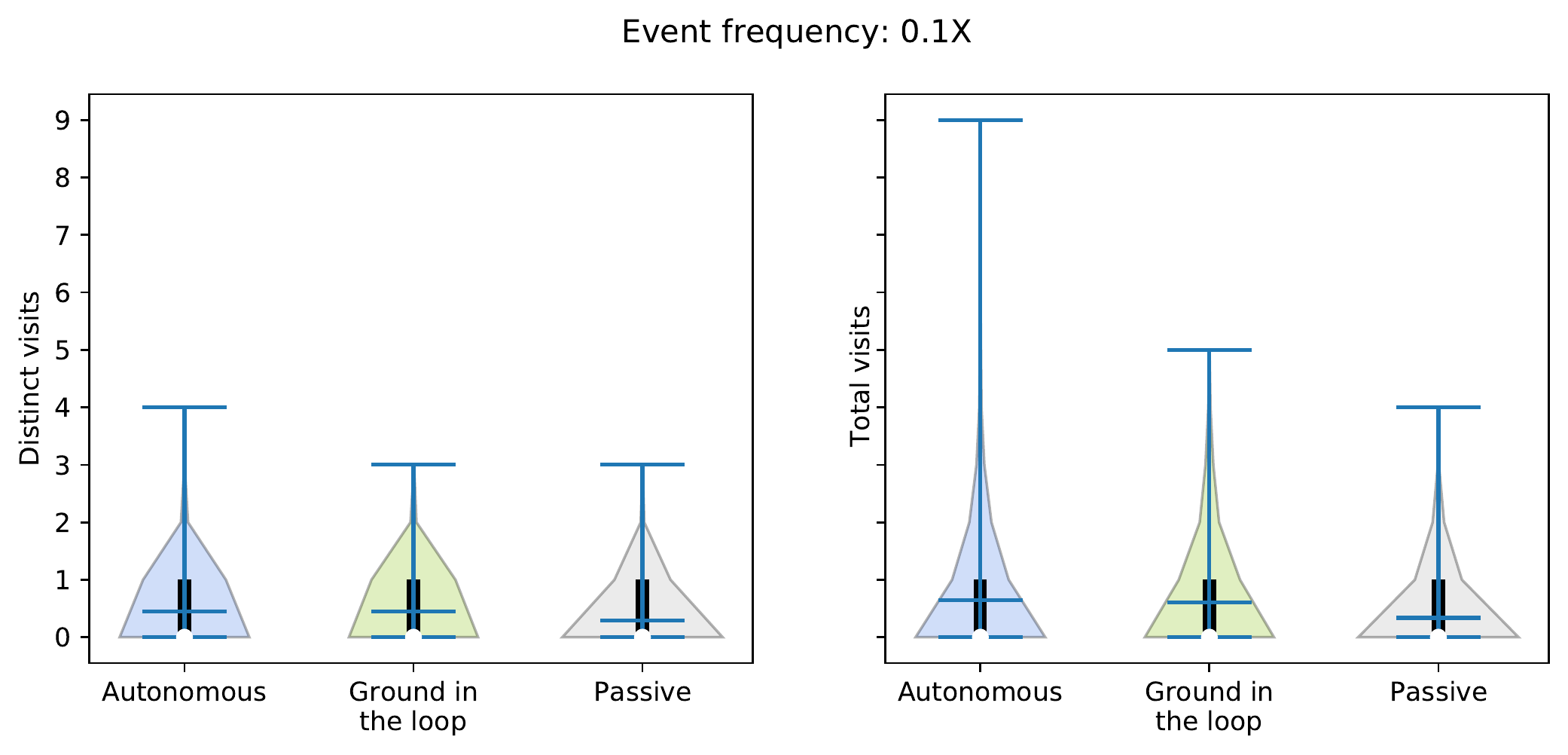}
\includegraphics[width=.7\textwidth]{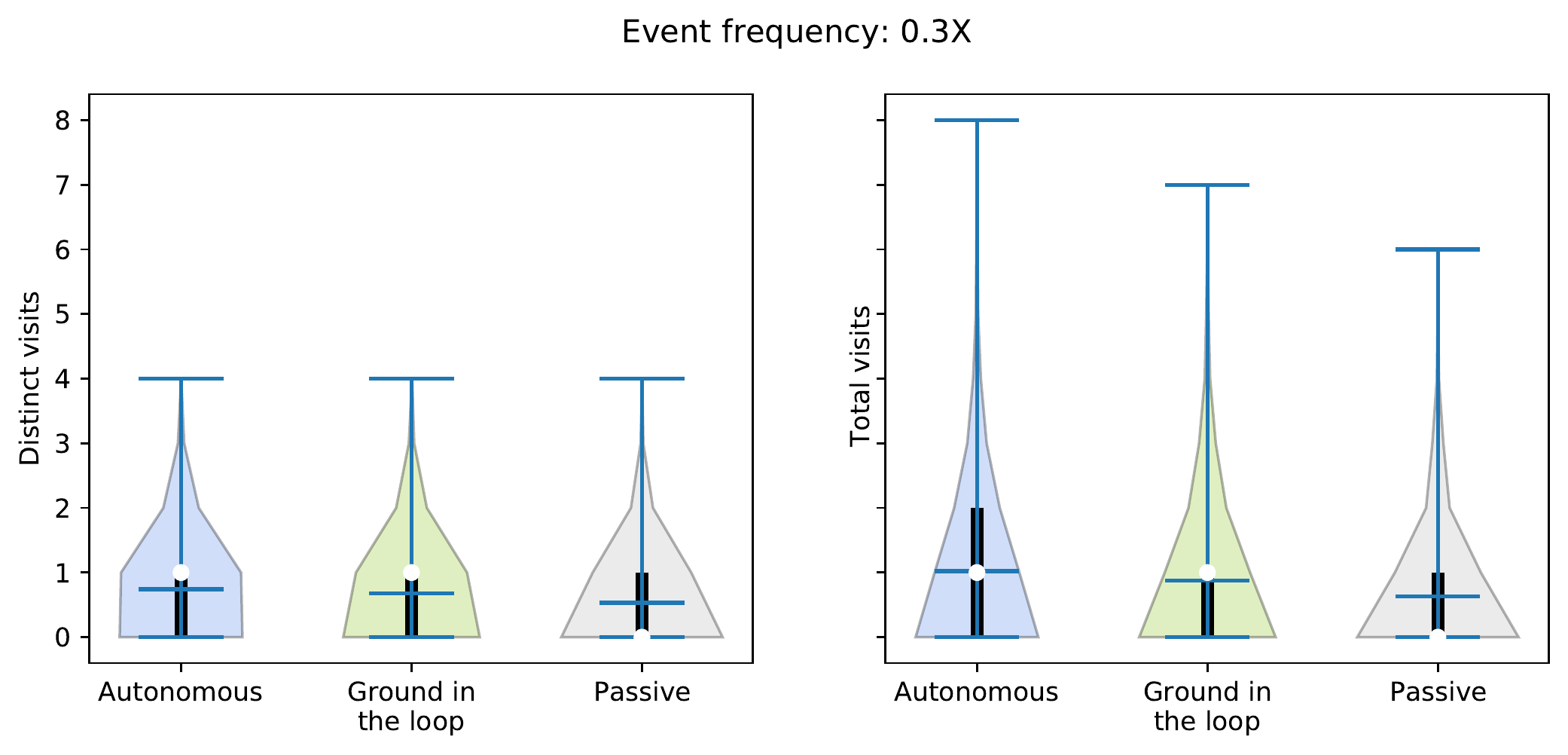}
\includegraphics[width=.7\textwidth]{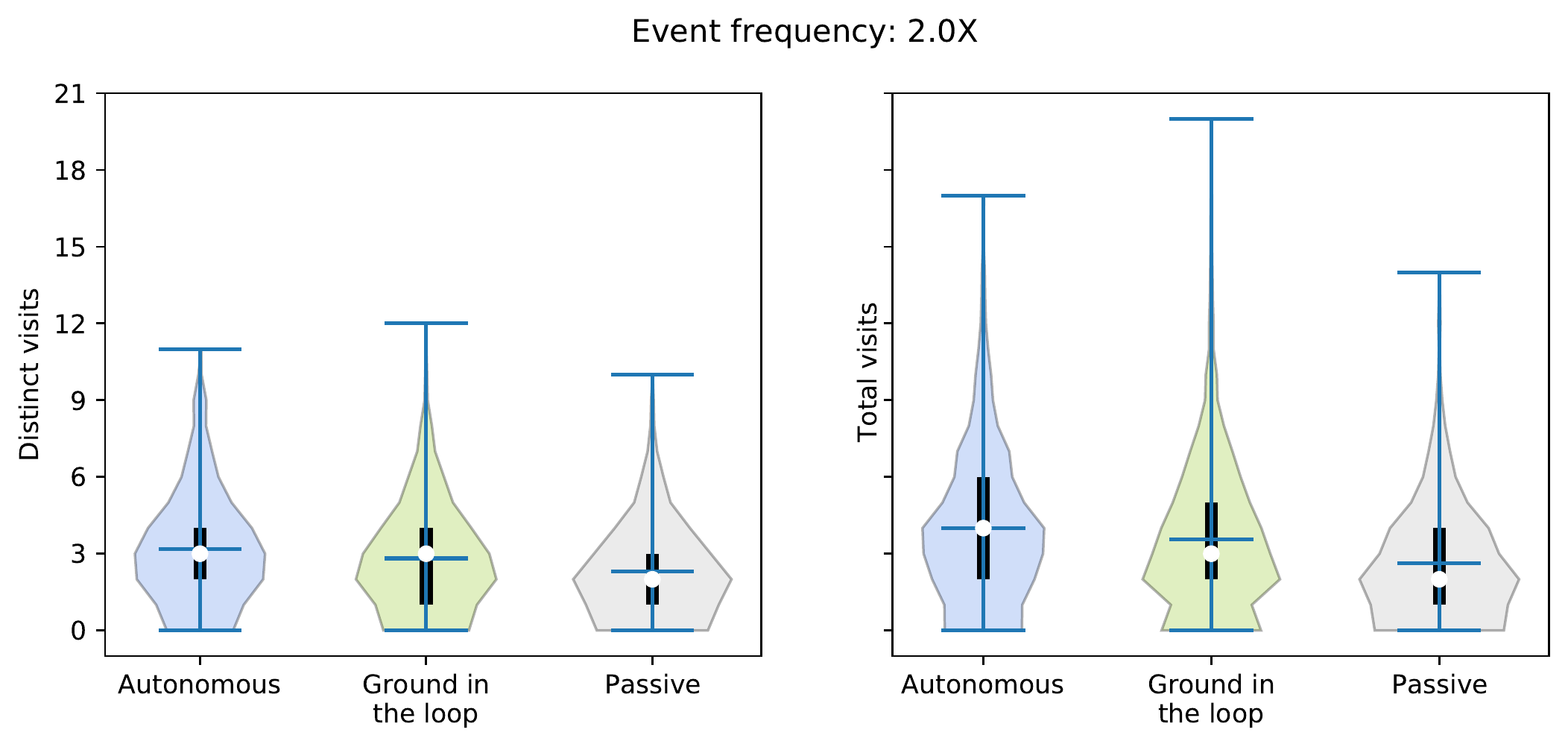}
\caption{Performance of the proposed approach compared with ground-in-the-loop and passive approaches with an event frequency of \revI{5\%}, 10\%, \revII{30\%,} and 200\% of the nominal one.}
\label{fig:results:event_frequency:comparison}
\end{figure}

Next, we assessed the sensitivity of the proposed approach to event frequency by considering scenarios where the likelihood of volcanic events occurring was 10\% and 200\% of the nominal ones, with a VAMOS orbiter.
For context, a recent study \cite{byrne2022estimate} %
estimated the frequency of eruptions on Venus as 73.9\% of Earth's, based on the planets' surface areas and masses.
Results are shown in Figures \ref{fig:results:event_frequency} and \ref{fig:results:event_frequency:comparison} and in Table \ref{tab:results:event_frequency:comparison}.

Critically, on-board autonomy maintains its advantage compared to passive approaches across a range of event frequencies (Figure \ref{fig:results:event_frequency:comparison}), yielding a mean 55\% increase in the number of distinct visits, and a  91\% increase in the number of total visits, when the event frequency is reduced tenfold. A 38\% increase in distinct visits and 51\% increase in total visits is achieved when the event frequency is double the nominal one. On-board autonomy also maintains a smaller, but significant advantage compared to ground-in-the-loop guidance for high event frequencies, with a 27\% increase in the number of distinct events visited and a 12\% increase in the number of total visits. Overall, these results confirm that the proposed approach is robust to large uncertainties in the event frequency, and holds particular promise in regimes when the overall number of observable events is low.

\revI{
\begin{figure}[h]
\centering
\includegraphics[width=\textwidth]{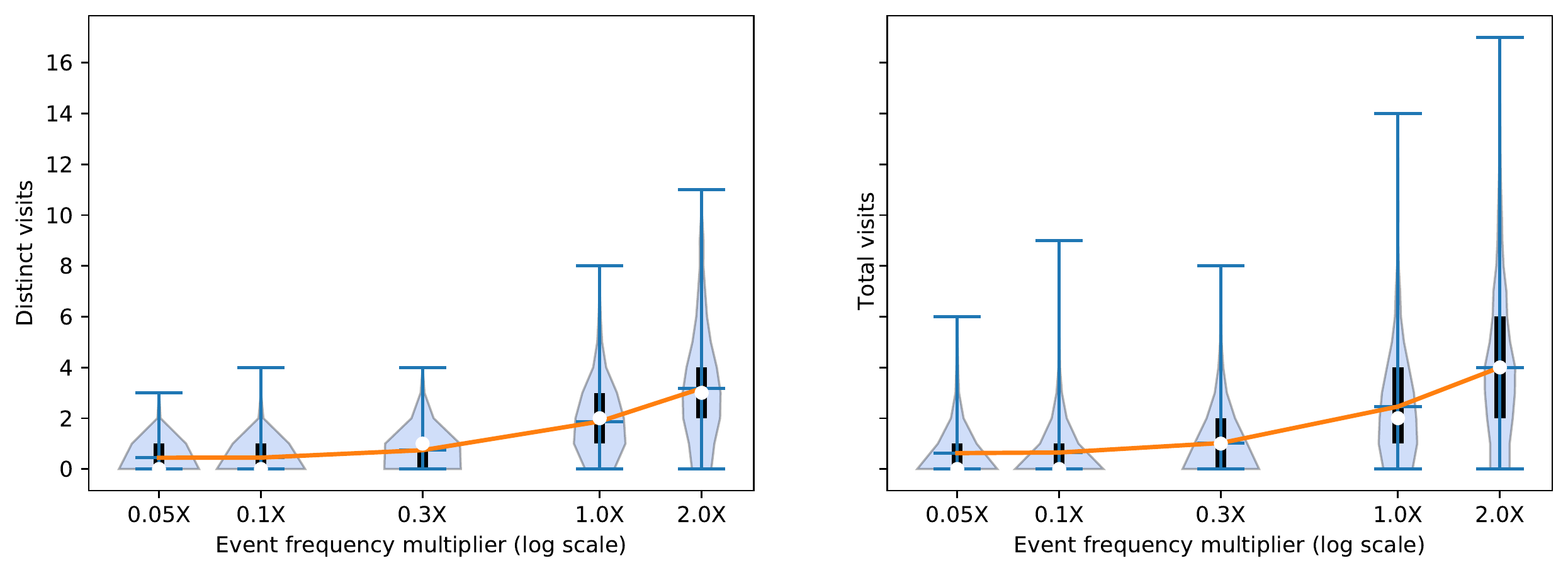}
\caption{Sensitivity of the approach's performance to the event frequency}
\label{fig:results:event_frequency}
\end{figure}
}

\begin{table}[h]
\caption{Performance of the proposed approach with an event frequency of \revI{5\%}, 10\%, \revII{30\%}, and 200\% of the nominal one. The nominal case is shown in Table \ref{tab:results:detections:VAMOS}.}
\label{tab:results:event_frequency:comparison}
\begin{adjustbox}{center}
\begin{tabular}{rlcccccc}
\toprule
Event &                               &  \multicolumn{2}{c}{Autonomous} &  \multicolumn{2}{c}{Ground in the loop} &  \multicolumn{2}{c}{Passive} \\
Frequency&                          &        mean &   std. dev. &                mean &           std. dev. &     mean &  std. dev. \\
\midrule

\multirow{5}{*}{\rotatebox[origin=c]{90}{0.05X}}&Distinct detections                &        \textbf{1.18} &        0.52 &                \textbf{1.18} &                0.51 &     \textbf{1.18} &       0.50 \\
&Distinct visits                    &        \textbf{0.44} &        0.55 &                0.38 &                0.52 &     0.29 &       0.47 \\
&Total visits                       &        \textbf{0.63} &        0.93 &                0.51 &                0.80 &     0.34 &       0.59 \\
&Detected events visited [\%]        &       \textbf{37.54} &       46.25 &               33.02 &               45.21 &    24.22 &      40.48 \\
&Events visited [\%]                 &       \textbf{33.62} &       43.30 &               29.18 &               41.71 &    21.46 &      37.31 \\

\midrule
\multirow{5}{*}{\rotatebox[origin=c]{90}{0.1X}}&Distinct detections                &        1.45 &        0.77 &                1.46 &                0.78 &     \textbf{1.47} &       0.77 \\
&Distinct visits                    &        \textbf{0.45} &        0.58 &                \textbf{0.45} &                0.58 &     0.29 &       0.50 \\
&Total visits                       &        \textbf{0.65} &        1.00 &                0.61 &                0.88 &     0.34 &       0.61 \\
&Detected events visited [\%]        &       \textbf{32.24} &       42.18 &               31.34 &               41.17 &    20.30 &      36.23 \\
&Events visited [\%]                 &       \textbf{26.51} &       37.15 &               25.51 &               35.76 &    17.02 &      32.04 \\

\midrule
\multirow{5}{*}{\rotatebox[origin=c]{90}{\revII{0.3X}}}&Distinct detections                &        \textbf{2.76} &        1.51 &                2.74 &                1.51 &     \textbf{2.76} &       1.51 \\
&Distinct visits                    &        \textbf{0.74} &        0.77 &                0.68 &                0.77 &     0.53 &       0.69 \\
&Total visits                       &        \textbf{1.02} &        1.21 &                0.87 &                1.09 &     0.63 &       0.89 \\
&Detected events visited [\%]        &       \textbf{28.71} &       31.92 &               26.65 &               32.26 &    21.07 &      29.78 \\
&Events visited [\%]                 &       \textbf{18.81} &       22.19 &               17.55 &               22.75 &    13.95 &      21.00 \\

\midrule
\multirow{5}{*}{\rotatebox[origin=c]{90}{2X}}&Distinct detections                &       15.33 &        6.21 &               15.36 &                6.25 &    \textbf{15.54} &       6.28 \\
&Distinct visits                    &        \textbf{3.18} &        2.19 &                2.82 &                2.04 &     2.31 &       1.80 \\
&Total visits                       &        \textbf{4.00} &        2.91 &                3.56 &                2.77 &     2.64 &       2.17 \\
&Detected events visited [\%]        &       \textbf{19.28} &       11.51 &               17.14 &               11.20 &    13.81 &       9.83 \\
&Events visited [\%]                 &       \textbf{11.66} &        7.24 &               10.36 &                7.06 &     8.46 &       6.21 \\
\bottomrule
\end{tabular}
\end{adjustbox}
\end{table}

\FloatBarrier
\subsection{Sensitivity to fleet size}

\begin{figure}[h]
\centering
\includegraphics[width=\textwidth]{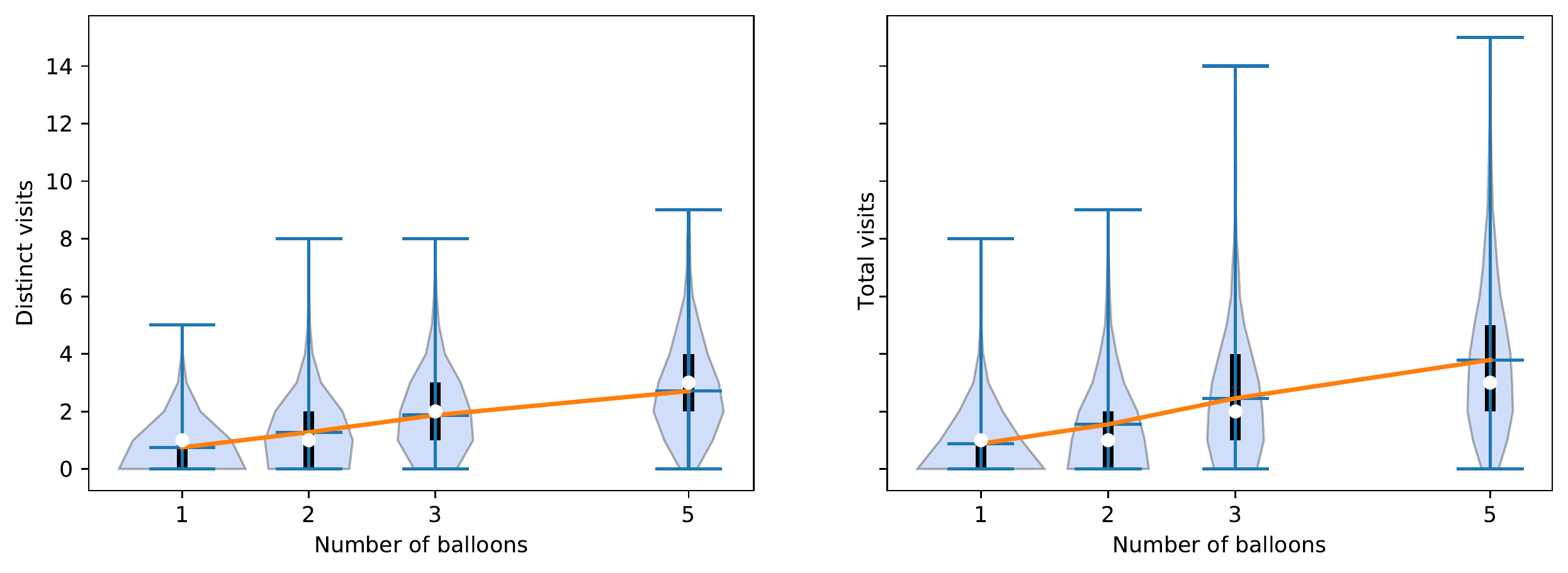}
\caption{Sensitivity of the approach's performance to the number of balloons}
\label{fig:results:num_balloons}
\end{figure}

Finally, we assess the performance of the proposed approach as the number of balloons changes from one to five, also with a VAMOS-type orbiter. Results are shown in Figure \ref{fig:results:num_balloons} and Table \ref{tab:results:num_balloons:all}. Unsurprisingly, the number of distinct and total visits grows approximately linearly with the number of balloons. More importantly, the advantage of on-board autonomy compared to the passive approach appears to hold for all fleet sizes, with increases in the number of total visits ranging from 51\% to 55\%, and increases in  \revII{distinct} events visited ranging from 34\% to 45\%, compared to the passive case. For small fleets, ground-in-the-loop guidance provides performance only slightly worse compared to on-board autonomy; however, for larger fleets, the gap between on-board autonomy and ground-based guidance widens, with an 8\% increase in total visits, and a 7.5\% increase in distinct events visited.

Collectively, these results show that the proposed approach holds promise to greatly increase the number of total and distinct events visited compared to no guidance; is highly robust to uncertainty in the detection radius and event frequency; and scales well across fleets of different sizes.

\begin{table}[h]
\caption{Performance comparison with one, two, and five balloons. The nominal three-balloon case is reported in Table \ref{tab:results:detections:VAMOS}.}
\label{tab:results:num_balloons:all}
\begin{adjustbox}{center}
\begin{tabular}{llcccccc}
\toprule
&                          &  \multicolumn{2}{c}{Autonomous} &  \multicolumn{2}{c}{Ground in the loop} &  \multicolumn{2}{c}{Passive} \\
                          &        mean &   std. dev. &                mean &           std. dev. &     mean &  std. dev. \\
\midrule
\multirow{5}{*}{\rotatebox[origin=c]{90}{One balloon}}&Distinct detections                &        \textbf{7.84} &        3.10 &                7.81 &                3.11 &     7.83 &       3.08 \\
&Distinct visits                    &        \textbf{0.75} &        0.85 &                0.71 &                0.84 &     0.53 &       0.74 \\
&Total visits                       &        \textbf{0.88} &        1.05 &                0.82 &                1.03 &     0.58 &       0.83 \\
&Detected events visited [\%]        &       \textbf{10.26} &       13.17 &                9.74 &               12.93 &     7.21 &      11.56 \\
&Events visited [\%]                 &        \textbf{5.62} &        7.13 &                5.40 &                7.16 &     4.07 &       6.67 \\

\midrule
\multirow{5}{*}{\rotatebox[origin=c]{90}{Two balloons}}&Distinct detections                &        7.75 &        3.07 &                7.76 &                3.10 &     \textbf{7.84} &       3.11 \\
&Distinct visits                    &        \textbf{1.28} &        1.18 &                1.19 &                1.10 &     0.88 &       0.95 \\
&Total visits                       &        \textbf{1.55} &        1.53 &                1.46 &                1.45 &     1.00 &       1.15 \\
&Detected events visited [\%]        &       \textbf{17.10} &       16.29 &               15.67 &               15.17 &    11.52 &      13.03 \\
&Events visited [\%]                 &        \textbf{9.75} &        9.38 &                9.26 &                9.44 &     6.81 &       8.19 \\
\midrule
\multirow{5}{*}{\rotatebox[origin=c]{90}{Five balloons}}&Distinct detections                &        8.54 &        3.38 &                8.54 &                3.40 &     \textbf{8.64} &       3.40 \\
&Distinct visits                    &        \textbf{2.71} &        1.64 &                2.52 &                1.72 &     2.01 &       1.53 \\
&Total visits                       &        \textbf{3.79} &        2.57 &                3.51 &                2.62 &     2.50 &       2.09 \\
&Detected events visited [\%]        &       \textbf{32.94} &       18.61 &               29.75 &               18.01 &    23.02 &      16.04 \\
&Events visited [\%]                 &       \textbf{20.52} &       12.22 &               18.75 &               11.81 &    14.73 &      10.77 \\
\bottomrule
\end{tabular}
\end{adjustbox}
\end{table}

\FloatBarrier
\section{Conclusions} %
\label{sec:conclusions}

In this paper, we explored how on-board autonomous event detection and guidance of buoyancy-controlled aerial platform can improve science returns for investigations of Venus's volcanic activity. We proposed a novel autonomy architecture comprising an on-board volcanic activity detector based on infrasound microbarometers; an autonomous guidance module that steers the balloons to targets of interest by exploiting Venus's uncertain wind circulation; and a shared event database that allows balloons and an orbiter to act as a team by sharing their knowledge of active volcanic events.%

Our results show that the proposed approach can yield an increase of up to 63\% in the number of follow-on observations compared to passive drifters; a smaller, but still significant, 40\% increase can be obtained if event detection is performed on board, but closed-loop guidance solutions for the balloons are computed on Earth.

We assessed the sensitivity of the approach to the choice of the orbiter's orbit, and showed that a \revI{high-altitude orbit that allows the orbiter to observe the full disc of the planet (e.g. the orbit of the proposed VAMOS mission)} offers increased performance compared to a lower orbit (e.g., the orbit of the VERITAS orbiter), due to increased opportunities for communication between balloons and the orbiter. Nevertheless, the proposed approach significantly outperforms passive guidance, irrespective of the selected orbit.

We also showed that the benefits of the proposed approach hold across a variety of detection radii and event frequencies, suggesting that the approach is robust to significant uncertainty in the underlying scientific phenomenon; and that the benefits hold for larger or smaller fleets of balloons, and are significant even for architectures with a single balloon.

A number of directions for future research are of interest.
First, we plan to extend the proposed autonomy approach to also optimize \emph{detection} of events of interest, guiding the balloons to maximize the likelihood of sensing previously-unobserved events whenever they are not actively steering towards an already-detected event.
Second, we will explore multi-agent reinforcement learning approaches where the balloons update the stochastic wind field model on the fly based on observations of their own movements, and use the updated model for planning.
Third, we will explicitly account for uncertainty in the balloons' knowledge of their location \revI{and orientation} through POMDP planning tools such as QMDP, similar to the approach in \cite{RossiBranchEtAl2021}.
Finally, and crucially, we plan to increase the technology readiness level of the proposed software architecture, with the end goal of a field demonstration of the full architecture, including on-board detection, on-board planning, inter-balloon communication, and active guidance, on hardware in a relevant Earth environment.

\section{Acknowledgement}
The research was carried out at the Jet Propulsion Laboratory, California Institute of Technology, under a contract with the National Aeronautics and Space Administration (80NM0018D0004).
\revI{The High Performance Computing resources used in this investigation were provided by funding from the JPL Information and Technology Solutions Directorate.} \revII{The authors thank Prof. Paul Byrne for the insightful comments and feedback.}

\bibliographystyle{unsrt}
\bibliography{references}

\end{document}